\documentclass{article}
\usepackage{iclr2027_conference,times}
\iclrfinalcopy
\usepackage{amsmath,amssymb,booktabs,tabularx,xcolor,graphicx}
\usepackage{tikz}
\usetikzlibrary{arrows.meta,calc}
\usepackage{hyperref,url}
\newcommand{\rev}[1]{{#1}}
\hypersetup{hidelinks,pdftitle={Collision-Free Is Not Enough: Passage-Conditioned Navigation for Wheeled-Legged Robots},pdfauthor={Wai Ka Sun, Waiyiu Pong, Bowen Wang}}
\title{Collision-Free Is Not Enough:\\Passage-Conditioned Navigation for\\Wheeled-Legged Robots}
\author{Wai Ka Sun$^{1}$, Waiyiu Pong$^{2}$ \& Bowen Wang$^{1}$\\
$^{1}$Columbia University\quad $^{2}$Tsinghua University\\
\texttt{sunwaikalithy@gmail.com}}
\begin{document}
\maketitle
\begin{figure}[h!]\vspace{-6pt}
\centering\includegraphics[width=\linewidth]{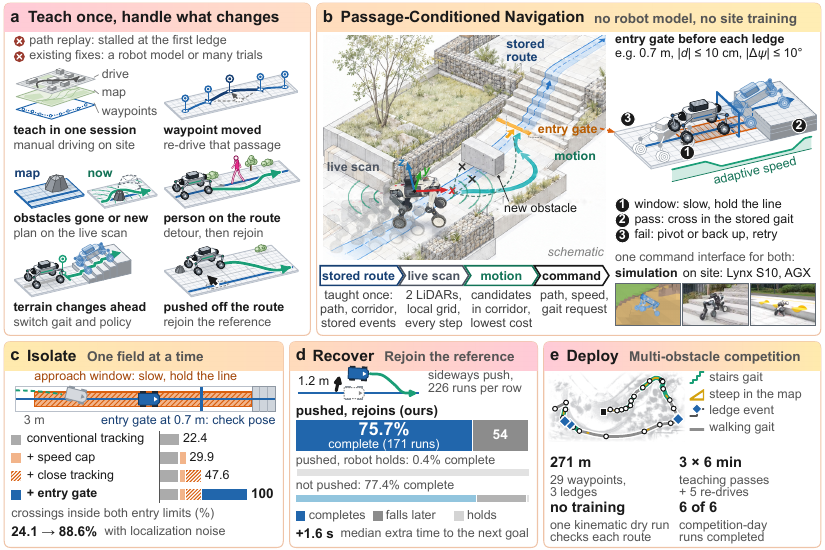}
\caption{\rev{\textbf{Path replay cannot adapt at transitions, and existing fixes need a robot model or many trials. We store how each transition was entered and enforce it after every replan.}
(a) One teaching session and five situations after it. (b) The stored route conditions motion on the live scan; an entry gate checks every ledge approach (schematic; simulation and site frames).
(c) \textsc{Isolate}: ledge crossings inside both entry limits as fields are added (Section~\ref{sec:isolate}). (d) \textsc{Recover}: outcomes after a 1.2\,m push (Section~\ref{sec:rejoin}). (e) \textsc{Deploy}: the competition route (Section~\ref{sec:physical}).}}
\label{fig:overview}
\end{figure}
\begin{abstract}
\rev{A collision-free detour can still bring a legged robot to the next stair or ledge a few centimetres or degrees off, and a fixed locomotion skill, factory or learned, can stall or fall. Existing methods prevent this with a robot model or extensive trial and error, but on a new site with unknown skills neither is available, so they cannot be used. We find that a single 27-minute teaching session already supplies what these methods have to derive or learn: for every stair and ledge, one crossing the skill has completed. Our method, passage-conditioned navigation, makes the route remember how each transition was entered, not only where the path went. It stores each taught crossing as an event and, without training or a robot model, enforces it after every local replan: before each ledge, an entry gate makes the robot turn or back up and retry if it is more than 10\,cm or $10^\circ$ off the taught crossing. Under localization noise, with path, gait schedule, and sensing unchanged, this raises the share of disturbed ledge crossings that stay within these limits from 24.1\% to 88.6\%. In contact simulation with our learned policies, a robot pushed off course before stairs never falls in 180 climbs, against 9 falls when the stored direction is ignored; pushed 1.2\,m away, it rejoins the route, ending on average 0.14\,m from it (1.10\,m without rejoining), and finishes as often as an undisturbed robot. This suits any robot with fixed skills that repeats a route over terrain transitions, as in inspection and patrol, where one wrong entry can end the task. We deployed it in competition on a real wheeled-legged robot: over 271\,m of multi-obstacle terrain spanning 7.0\,m of elevation, it completed all six runs, won third place at the 2026 Global Open-source AI Challenge, which drew 2,999 submissions, and was the only finalist in its task to finish fully autonomously.}
\end{abstract}

\section{Introduction}
\leavevmode\rev{Wheeled-legged robots can take over outdoor work that is dangerous for people, such as
shipyard and industrial inspection, mountain patrol, and search and rescue. Their routes chain stairs and ledges with loose ground and slopes, and the robot has to cross the same transitions on every run. We met the difficulty while preparing such a route for a competition. An earlier version of our system simply followed the taught path and treated ledges as ordinary waypoints, and it stalled at the first ledge in both full-speed runs: the robot's ledge gait mounts a ledge only when the robot arrives within about ten centimetres and ten degrees of a workable entry, and nothing in the path enforced this. Avoiding collisions is not enough either. After a detour, the robot may have too little room to slow down, may need a
gait change before the controller is ready, or may reach the next step at an angle. Every successful crossing, however, shows an entry that works, and this experience is worth keeping.}

\rev{Keeping it is hard: replaying the taught track cannot handle local changes, and editing its geometry can throw away what made the next passage work. This leads to our question:}
\emph{how can a robot adapt to local changes \rev{and still keep the entry conditions that
a successful passage showed?}}

\rev{Locomotion on legged robots often comes as a fixed library of skills: gaits that ship with the robot, as on our platform, or learned policies that a higher-level module chooses among \citep{hoeller2024anymal,wu2025vocaloco}. Each skill succeeds only from a limited set of entry states, its \emph{initiation set} in
the language of options \citep{sutton1999between}, and a detour can leave the robot outside
that set where skills switch at a stair or ledge. Existing methods obtain this set in one of two ways: planners derive entry constraints from a robot model, and skill-chaining methods learn initiation sets from many further trials (Section~\ref{sec:related}). For a frozen library on a new site, neither way is available: the skills are black boxes, and controllers that come with the platform or were trained once cannot be retrained for each site. We find that one teaching session already supplies what these methods derive or learn: for every transition, one crossing that the skill has completed. We therefore treat the library as a black box and retrain nothing. On the robot the skills are the factory gaits; in simulation they are policies that we trained with PPO and then froze. One teaching session in the
real environment (three passes of about six minutes and five short re-drives, 27 minutes of
driving in all) crosses every transition. Building the stored route needs no training, and
autonomous runs began the same day.}

\rev{We store each terrain transition on the route as an event, placed by its distance along
the route, so that it persists across runs and still applies after a local replan. Each stairs
or ledge event carries a gait interval and a speed cap. A ledge event also carries the
position and direction of the taught crossing (its \emph{crossing frame}) and an
\emph{approach window}, inside which the robot slows down and steers toward a point close
ahead. At a fixed distance before the ledge, an \emph{entry gate} checks the robot's sideways
offset and heading in the crossing frame; if the check fails, the robot turns in place or
backs up and approaches again (Figure~\ref{fig:method}). Teaching supplies each event's
position, direction, and gait interval; the limits of the check (0.10\,m, $10^\circ$) and the
other settings are shared by the whole environment. In the language of options, the gate
tests at run time whether the robot is inside a conservative part of the next skill's
initiation set, centred on the taught crossing. The method is simple and cheap, and it adapts
during the run: the robot detours around obstacles that are not on the map, corrects its
pose before each ledge, and returns to the route within 2\,s of a stall.}

\paragraph{\rev{Contributions.}} \rev{(i) A route representation that keeps what teaching
showed (position, direction, gait interval) apart from the rules we configure, and ties each
stored item to the run-time decision it changes (Table~\ref{tab:fielddecision}).
(ii) A controlled test of what the stored events add, with path, gait schedule, and sensing
fixed. Conventional tracking brings 22.4\% of disturbed ledge crossings within the entry
limits, the approach window raises this to 47.6\%, and the entry gate to 100\% (24.1\%,
46.0\%, and 88.6\% under localization noise). Of 1,470 approaches made with the window but without the gate, all
350 that pass the gate's check end within the limits. With our learned policies in contact
simulation, a robot pushed off course before stairs does not fall once in 180 climbs, against
9 falls when the stored direction is ignored, and ledge entries outside the gate's limits fall more than twice as often as entries
inside (Section~\ref{sec:isolate}).
(iii) Because the stored route persists, a robot pushed 1.2\,m off it returns and completes
75.7\% of segments, the level of runs without a push (77.4\%), in contact simulation over 23
seeds (Section~\ref{sec:rejoin}). On the robot, with factory gaits, the system completed all
six competition-day runs in a complete multi-obstacle environment and won third place (Section~\ref{sec:physical}).}

\section{Related work}\label{sec:related}
\paragraph{Teaching, route reuse, and replanning.}
\rev{Teach-Repeat-Replan turns imperfect demonstrated flight trajectories into smooth
references that the vehicle can follow, and replans around unmapped or moving obstacles
\citep{gao2019teach}. LiDAR teach-and-repeat on difficult ground \citep{baril2022kilometer},
adaptive route following \citep{sehn2022lines}, and trajectory-based predictive control
\citep{Rosolia_2020} develop route reuse and adaptation further. \citet{sehn2022lines} plan in
a curvilinear space and penalize deviation from the taught path to keep what is known about
the terrain. Building on route-relative planning, we store entry geometry and gait intervals
for wheeled-legged terrain transitions and enforce them at run time. Commercial
teach-and-repeat navigation for legged robots likewise marks the edges of a recorded route
with hints for the locomotion controller, for example that an edge crosses stairs or needs
alignment \citep{bostondynamics2026graphnav}. These services are closed-source, and we know of
no published account or evaluation of how they enforce entry; ours is released. A penalty on
path deviation keeps the path. We also keep discrete events (a crossing frame, an approach
window, a gait interval) and enforce each one where it occurs.}

\paragraph{Wheeled-legged planning and locomotion.}
\rev{Whole-body MPC with online gait generation \citep{bjelonic2021wholebody} and motion
libraries with online MPC \citep{bjelonic2022offline} coordinate mobility and feedback.
\citet{lee2024robust} combine learned locomotion, mobility-aware local navigation, and
large-scale planning, and deploy them over kilometres in cities. Their navigation and
locomotion controllers are both learned and are joined through hierarchical reinforcement
learning into one learned stack. We study how to keep task-specific passage conditions while
reusing an existing gait library. Our interface sits on a frozen skill library and trains
nothing: factory gaits on the robot, frozen policies in simulation. It also makes route preparation and execution decisions explicit.}

\paragraph{\rev{Kinodynamic and manoeuvre-constrained planning.}}
\rev{Kinodynamic planning \citep{donald1993kinodynamic,lavalle2001randomized} and
manoeuvre-constrained replanning derive the approach pose, speed, and mode from a model of the
system. A frozen black-box skill library provides no such model. We take the constraint from
one taught crossing per transition and enforce it at run time as a test of a conservative part
of the next skill's initiation set. In our logs, every approach that the test accepts succeeds
(Section~\ref{sec:isolate}).}

\paragraph{\rev{Skill composition and initiation sets.}}
\rev{In the options framework, a skill has an initiation set \citep{sutton1999between}, and
skill chaining learns these sets so that one skill ends where the next can begin
\citep{konidaris2009skill,bagaria2020option}. They learn initiation classifiers from repeated interaction. Related work shapes where each policy ends during training \citep{lee2022adversarial}, scores a skill sequence with the skills' value functions \citep{agia2023stap}, and inserts motion plans that bring the robot to a configuration from which a black-box skill can start \citep{hedegaard2025task}. Other work trains a navigation policy over locomotion skills \citep{hoeller2024anymal} or predicts from perception whether each pretrained
locomotion policy will succeed \citep{wu2025vocaloco}. Path-conditioned planning and
policy-dependent traversability consider reference guidance and controller capability
\citep{haro2026path,Yan_2026}, and sequential composition orders feedback controllers so that
the goal of one lies in the domain of the next \citep{burridge1999sequential}. We work with a
frozen library, one teaching session, and no further interaction. The stored event records
where and in which direction the next skill began in teaching; the gate enforces a tolerance around it.}

\section{Task and observation interfaces}
\leavevmode\rev{The inputs are a point-cloud map $M_0$, ordered goal regions $W$, an
existing locomotion library $\mathcal G$, and $K$ real teaching episodes $\mathcal D_H^{(K)}$.
A reference $\Pi^0=(\gamma^0(s),v^0(s),g^0(s))$ gives geometry, speed limits, and gait
intervals as functions of the route coordinate $s$. Compilation and execution use explicit
rules and geometric operations, and the executor schedules controllers from the library.}

\begin{figure}[t]
\centering\includegraphics[width=\linewidth]{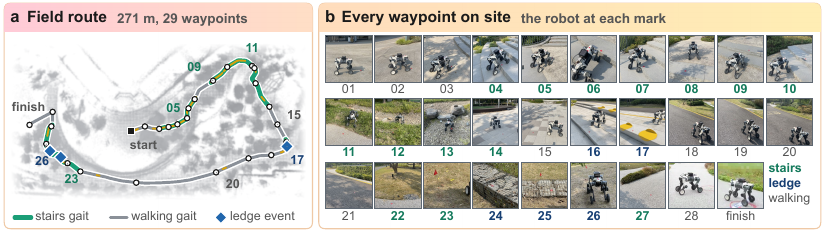}
\caption{\rev{\textbf{The field route, and the robot at every waypoint.} (a) The stored route on the field map, coloured by stored gait, with its 29 waypoints and three ledge events. (b)
The robot at each waypoint on site, numbered in route order; each number takes the colour of the route at that waypoint.}}
\label{fig:fieldmap}
\end{figure}

\rev{The compiler reads pose trails, waypoint marks, gait events, and map geometry. Teaching
supplies route geometry and gait events. We configure speed limits, the choice of demonstrations, corridor widths, approach extents, and geometric thresholds. In the field, a goal is
reached when the robot's body touches its $0.25\,\mathrm m$ waypoint disk; simulation requires
the robot centre to come within $0.20\,\mathrm m$ of each goal, in order. Goal positions and
their order are fixed within each evaluation (Appendix~\ref{app:goalinterfaces}). The task and
the stored events share the coordinate frame of the field point cloud
(Figure~\ref{fig:fieldmap}), so the executor switches local conditions on by route progress.}

\section{Method: passage-conditioned navigation}\label{sec:method}

\begin{figure}[t]
\centering\includegraphics[width=\linewidth]{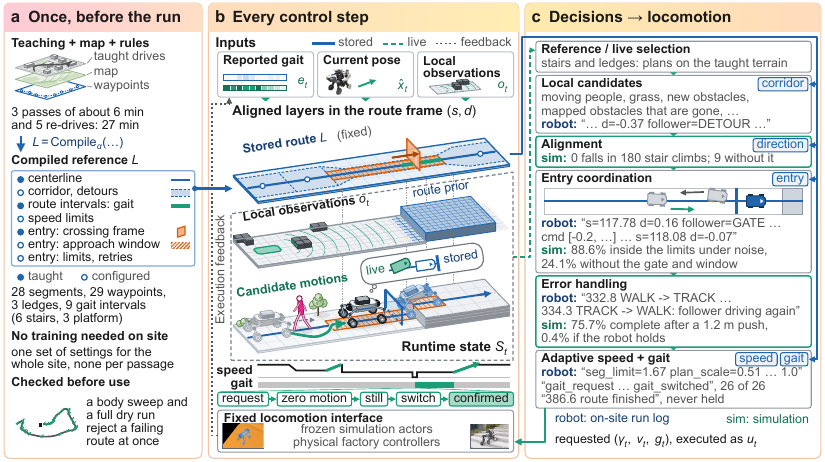}
\caption{\rev{\textbf{Passage-conditioned navigation: what is stored, what is observed, and what each decision does.} (a)~Teaching, the map,
and configured rules compile once into the stored route $L$; a body sweep and a full kinematic run check it before
use. (b)~At every control step, the stored route, the local observations, and the candidate motions share one route
frame; the robot compares the stored entry with its live pose and adapts its speed along the passage. (c)~The
decisions of one step, with excerpts of the robot's log and simulation results.}}
\label{fig:method}
\end{figure}

\rev{\paragraph{Overview.} The reference stores each terrain transition as an event, placed by
its distance along the route (Table~\ref{tab:fielddecision}). Every stairs or ledge event has a
gait interval and a speed cap; each ledge event also has a crossing frame and an approach
window. A local replan may change the path, but never the events. Inside the window the robot
slows down and tracks the stored line closely. 0.7\,m before the crossing, the entry gate
checks the pose in the stored frame (Eq.~(\ref{eq:gate})); if the check fails, the robot turns
in place or backs up and approaches again. Sections~\ref{sec:compile} to~\ref{sec:enforce}
describe how the route is compiled, repaired, and enforced. Figure~\ref{fig:method} shows what is stored and what is observed.}

\subsection{Compiling a persistent reference}\label{sec:compile}
The compiler returns a reference package
\begin{equation}
L=\operatorname{Compile}_{\alpha}(M_0,W,\mathcal D_H^{(K)})
=(M_0,W,\Pi^0,\mathcal K_E),
\label{eq:compile}
\end{equation}
\rev{where $\alpha$ holds the explicit rule settings and $\mathcal K_E$ stores the corridor,
gait, approach, and switch conditions. The locomotion library $\mathcal G$ is paired with this package for validation and execution.}

\rev{The compiler aligns the waypoint marks with the selected pose trails and uses the
waypoints as anchors for correction intervals. The joined polyline is resampled by cumulative
planar arc length, $s_i=s_{i-1}+\|p_i-p_{i-1}\|_2$, and aligned to the anchors.
Corridor, body-sweep, and clearance checks decide whether a shortcut or a rounded join is
allowed; otherwise the taught geometry is kept.}

\rev{The \emph{platform gait} is the library's gait for mounting a ledge. Where the teaching
trail uses it, the compiler records a ledge crossing. Its position and approach direction come
from the trail and can be adjusted to the ground profile in the point cloud. Gait events from
teaching are projected onto the joined line and become gait intervals along the route. For
each segment, the package writes the centerline, gait, speed limit, detour permission, and
corridor width, together with the crossing and approach fields. The crossing position and direction therefore come from the demonstration; we configure the approach extents, tolerances, speed caps, and retry rules (Appendix~\ref{app:implementation}). Because everything shares one route coordinate, a later local decision can refer to a taught transition.}

\rev{A \emph{route-prior observation} is a height map built from the elevation of the stored
route and used in place of the live grid. Because the passage was recorded, it can override an
overly cautious map classification, under the recorded speed and approach conditions.}

\subsection{Repairing the route with short-lived state}\label{sec:repair}
\leavevmode\rev{Let $o_t$ be the local observations that the planner uses, $\hat x_t$ the localized pose, $e_t$ the
execution reports, and $S_t$ the recent state. The run-time interfaces are}
\begin{align}
S_{t+1}&=\operatorname{Update}(S_t,o_t,\hat x_t,e_t),\label{eq:update}\\
(\gamma_t,v_t^{\rm req},g_t^{\rm req})&=\operatorname{Adapt}(L,S_{t+1}),\label{eq:adapt}\\
u_t&=\operatorname{Execute}(\gamma_t,v_t^{\rm req},g_t^{\rm req},\hat x_t,g_t^{\rm confirmed}).
\label{eq:execute}
\end{align}
\rev{$S_t$ holds recent observations, route progress, tracking error, and timers; a stale
observation stops the follower. The stored route $L$ stays fixed during a run; it changes only
when recompiled and reloaded.}

\rev{On live-sensing segments, the pose is projected into route-relative coordinates $(s,d)$.
Candidate curves are generated inside the current corridor, each with a smooth (quintic)
sideways shift and a return toward the next goal. Each candidate's footprint is swept over the
local grid. For a candidate offset $d_j$, the planner uses the cost}
\begin{equation}
J_j=w_o\rho_j+w_f(1-f_j)+w_l|d_j|+w_s|d_j-d_{\rm prev}|
-w_p f_j+w_b\mathbf1[d_jd_{\rm prev}<0],
\label{eq:cost}
\end{equation}
\rev{where $\rho_j$ is the obstacle proximity and $f_j$ the fraction of the candidate that is
free. The planner rejects candidates whose free stretch at the start is too short. Where the stored route permits no detour, as on stairs, the only candidate is the stored line; when the next goal lies within the 3.5\,m planning horizon, every candidate returns to zero offset within the last metre before it. A switching
margin keeps the current candidate, and a bounded A* search is the fallback where detours are
permitted. The stored route defines the candidate coordinates and the permitted corridor; local
evidence changes which candidates are valid, their cost, and the speed.}

\begin{table}[t]
\centering\small
\begin{tabularx}{\linewidth}{@{}p{0.30\linewidth}X@{}}
\toprule
Stored field & Runtime decision it changes\\
\midrule
Centerline & Defines the candidate coordinates $(s,d)$; every detour is expressed relative to the stored passage location\\
Corridor, detour permission & Bound the candidate offsets $d_j$ and gate the A* fallback\\
Speed cap, approach window & Scale the command and start the slow, short-lookahead approach before a ledge\\
Gait interval & Fixes where a mode request is issued and held\\
Crossing position and direction & Define the frame of the entry gate; a failed check triggers a pivot or a back-up and a new approach\\
\bottomrule
\end{tabularx}
\caption{\rev{\textbf{Each stored field and the decision it changes.} Eq.~(\ref{eq:cost}) scores clearance, route-relative offset, and switching stability; entry
conditions act through the approach window and the entry gate.}}
\label{tab:fielddecision}
\end{table}
\subsection{Speed, gait, and entry at run time}\label{sec:enforce}
\leavevmode\rev{The follower keeps commands within the segment's speed limit, slows down near obstacles,
and requests the segment's gait. A gait request is not yet a gait change: the handoff asks for
zero motion, waits until the robot is still, switches, and waits for confirmation (Appendices~\ref{app:actorinterface} and~\ref{app:goalinterfaces}).}

\paragraph{\rev{Approach window and entry gate.}}
\rev{A platform-gait interval keeps a crossing position $p_j$ and direction $\psi_j$. They
define a local approach frame:}
\begin{equation}
\begin{bmatrix}\ell_{j,t}\\d_{j,t}\end{bmatrix}
=R(-\psi_j)(p_t-p_j),\qquad
\Delta\psi_{j,t}=\operatorname{wrap}(\psi_t-\psi_j).
\label{eq:entryframe}
\end{equation}
\rev{Two stored quantities act on entry. The \emph{approach window} is the stretch of route
$[s_j-b_j-1\,\mathrm m,\ s_j+a_j]$, where $b_j$ and $a_j$ are the lengths of the straight
crossing lane before and after the crossing at $s_j$. The lane is set to $2.0\,\mathrm m$
before and $1.0\,\mathrm m$ after the crossing, and is shortened where clearance or a waypoint
requires. Inside the window, the forward command is capped at the approach speed and the robot
steers toward a point $0.5\,\mathrm m$ ahead; elsewhere the follower looks at least
$1.0\,\mathrm m$ ahead. The \emph{entry gate} checks the pose once, at $\ell_{j,t}=-\ell_c$
with $\ell_c=0.7\,\mathrm m$:
\begin{equation}
G_j=\mathbf 1\!\left[\,|d_{j,t}|\le\delta_d\ \wedge\ |\Delta\psi_{j,t}|\le\delta_\psi\right],
\qquad \delta_d=0.10\,\mathrm m,\ \delta_\psi=10^\circ .
\label{eq:gate}
\end{equation}
With $n$ the number of back-ups already used at this ledge, the executor then
\begin{equation}
\begin{cases}
\text{crosses} & G_j=1,\\
\text{pivots toward }\psi_j\text{ and checks again} & |d_{j,t}|\le\delta_d,\ |\Delta\psi_{j,t}|>\delta_\psi,\\
\text{backs up }0.3\,\mathrm m\text{ and approaches again} & |d_{j,t}|>\delta_d,\ n<R,\\
\text{crosses} & n=R,\ \delta_d<|d_{j,t}|\le\delta_r,\\
\text{holds} & \text{otherwise},
\end{cases}
\label{eq:gatepolicy}
\end{equation}
with $R=2$ and a relaxed bound $\delta_r=0.25\,\mathrm m$. The crossing frame $(p_j,\psi_j)$
comes from teaching; $\delta_d$, $\delta_\psi$, $\ell_c$, $R$, and $\delta_r$ are settings
shared by all ledges. Because the check uses the stored crossing frame, it
still applies after a local route change; a robot that starts beyond the check point is not
checked. Section~\ref{sec:isolate} separates the window from the gate, and
Appendix~\ref{app:isolation} varies $R$ and $\delta_r$.}

\paragraph{\rev{What the planner sees.}}
\rev{On live-sensing segments, the planner uses the local grid and scan from the LiDAR. On
selected stairs and platform intervals and around ledge crossings, the field configuration
uses route-prior observations instead. This lets the robot cross familiar terrain that the
flat-ground grid would reject, and leaves obstacle response in those intervals to the route
prior. Live obstacle sensing is active on 58\% of the field route; there, the local planner
detours around obstacles that are not on the map. The substitution is limited to stairs and
ledge intervals, which the robot crosses inside a narrow corridor, and to the
\emph{route-prior window}, the $1.5\,\mathrm m$ that follow the platform-gait interval of each
ledge. A robot that stalls or leaves the line \emph{rejoins} the reference: it walks forward
while steering to a point ahead on the stored line. As a safety fallback, it holds after
$20\,\mathrm s$ of rejoining without $0.5\,\mathrm m$ of progress, or when it is
$1.5\,\mathrm m$ off the line.}

\begin{figure}[t]
\centering
\includegraphics[width=\linewidth]{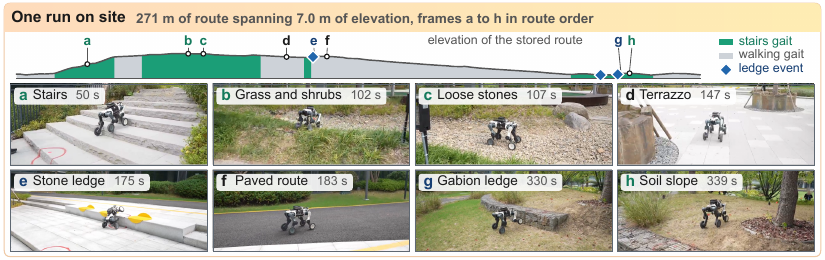}
\caption{\textbf{\rev{One run on the robot across successive terrain transitions.}}
\rev{Eight frames in route order, each with its time in the run. The strip plots the elevation of the stored
route, coloured by stored gait, and marks where each frame was taken. Bystanders' faces are pixelated.}}
\label{fig:physicaloverview}
\end{figure}

\section{Evaluation}
\leavevmode\rev{We ask four questions. With geometry, gait schedule, and sensing held fixed, what do the
stored fields add (Section~\ref{sec:isolate})? Does returning to the stored route recover runs
after a push (Section~\ref{sec:rejoin})? Where do route preparation and the choice of
observations help (Section~\ref{sec:routeprep})? Does the system hold up on the real robot in a
complete competition environment (Section~\ref{sec:physical})? Entry geometry is measured in kinematic simulation on the route
compiled from teaching in the complete multi-obstacle competition environment. Recovery and
route preparation use contact simulation on a scene reconstructed from the map, with its own
stored route. The end-to-end result comes from the robot's logs and the competition record.}

\paragraph{Simulation protocol.}
\rev{The MuJoCo navigation core uses two frozen policies, \texttt{J3100.onnx} and
\texttt{stairs\_1150.onnx}, which differ from the factory gaits on the robot. We trained both
with PPO and froze them; no experiment here retrains them. The simulation routes are adapted
from a draft built from the mapping trajectory on the reconstructed scene. Matched comparisons
share policies, goals, sensors, and condition parameters. The pose-noise condition adds
Gaussian noise with standard deviations of 0.03 m in planar position and $1^\circ$ in yaw.
Unless stated otherwise, a run succeeds when the robot centre passes within 0.20 m of every
goal in order; falls, blocked states, and timeouts are kept in the reported outcomes.}

\subsection{Teaching and compilation}
\leavevmode\rev{Teaching is one session in the real environment: the operator drives the route two or
three times, about six minutes per pass. Here three passes took 6.1, 6.1, and 5.9 minutes,
and the last two also supply the gait intervals. Where a passage needs it, a re-drive of
that passage is spliced in. Five are used here, of 0.7 to 4.6 minutes; the longest covers
111 m of the route.  After the splices are chosen and the geometry is processed, the compiled
route has 270.9 m of centerline spanning 7.0 m of elevation, 29 waypoints, and 28 segments. Each ledge is crossed in one
of the spliced re-drives, so its crossing position and direction come from one recorded
traversal.  Before it is
installed, every candidate route must pass a body-sweep clearance check against the map and a
kinematic run of the full route; during preparation these two checks rejected candidates 24
times (Appendix~\ref{app:preparation}; Figure~\ref{fig:teaching}). Planning from the map and
waypoints alone gave no usable route: the better of two grid planners found a path for 13 of
29 segments (over the 30 marks of that stage; Appendix~\ref{app:preparation}).}

\subsection{What the stored fields add}\label{sec:isolate}
\leavevmode\rev{We run the field navigation core on this route in a deterministic kinematic simulation.
Before each of the three stored ledges we inject a sideways offset of $0$, $\pm0.1$, $\pm0.3$,
or $\pm0.4$ m combined with a heading offset of $0$, $\pm7.5^\circ$, $\pm22.5^\circ$, or
$\pm30^\circ$: 49 cells and 147 ledge records per arm. The route geometry, gait plan,
observations, and command and feedback models are the same in every arm. We score the share
of ledge crossings within both entry limits of Eq.~(\ref{eq:gate}).}

\paragraph{\rev{Separating the mechanisms.}} \rev{Ledge handling combines four mechanisms,
which the kinematic diagnostic of Appendix~\ref{app:kinematic} switches on or off together. We give each its own switch: the
approach-speed cap, close tracking over the approach window, the entry gate, and the
route-prior window after the ledge. We ran the full $2^4$ factorial (784 rollouts, 2,352
crossing records) with the offset injected $0.76\,\mathrm m$ before the ledge. The all-off arm
is conventional tangent tracking with gait scheduling on the same centerline. It and the
all-on arm reproduce the two arms of that diagnostic exactly.}

\begin{table}[t]
\centering\small
\begin{tabular}{@{}lrrr@{}}
\toprule
Arm (offset injected 0.76 m before the ledge) & Inside both limits & Mean $|d|$ & Mean $|\Delta\psi|$\\
\midrule
Conventional tracking and gait schedule & 33/147 = 22.4\% & 13.43 cm & $22.30^\circ$\\
+ approach-speed cap & 44/147 = 29.9\% & 8.82 cm & $20.00^\circ$\\
+ short-lookahead tracking (full approach window) & 70/147 = 47.6\% & 5.81 cm & $14.43^\circ$\\
+ entry gate & 147/147 = 100\% & 0.57 cm & $1.54^\circ$\\
+ route-prior window after the ledge & 147/147 = 100\% & 0.56 cm & $1.55^\circ$\\
\midrule
Entry gate alone & 138/147 = 93.9\% & 1.13 cm & $3.52^\circ$\\
\bottomrule
\end{tabular}
\caption{\rev{\textbf{The stored fields, added one at a time.} Geometry, gait plan, and
sensing are identical across arms. Main effects over the other eight
combinations: entry gate $+64.5$ points (simple effects $+52.4$ to $+71.4$), speed cap $+8.3$,
tracking window $+7.3$, route-prior window $-0.2$ (Appendix~\ref{app:isolation}).}}
\label{tab:ladder}
\end{table}

\rev{Table~\ref{tab:ladder} adds the fields one at a time, starting from conventional
tracking. Conventional tracking with the speed cap and the close tracking of the approach window reaches 47.6\%. The gate supplies the rest and reaches 93.9\% alone. At these events the stored heading agrees with the direction of the local
centerline to within $0.01^\circ$ (Appendix~\ref{app:kinematic}), so every arm is given the
same direction. What the event adds is the check: where the pose is tested, against which
frame and limits, and what happens when the test fails. The route-prior window leaves entry
geometry unchanged ($-0.2$ points).}

\paragraph{\rev{Dependence on disturbance distance.}} \rev{We repeated three of these arms with the offset injected earlier (Table~\ref{tab:distance}). When the offset arrives
$1.2\,\mathrm m$ or more before the ledge, the approach window alone brings every crossing
within the limits; at $1.2\,\mathrm m$ the gate ended one of 49 rollouts in a hold.
Conventional tracking needs about $2.4\,\mathrm m$ of undisturbed approach to come close (95.9\%).
The two event fields split the work: the window handles disturbances that arrive early, and
the gate handles those that arrive just before its check point, as after a detour that
rejoins the route late.}

\begin{table}[t]
\centering\small
\begin{tabular}{@{}lrrr@{}}
\toprule
Offset injected & Conventional tracking & + approach window & + entry gate\\
\midrule
2.4 m before the ledge & 95.9\% & 100\% & 100\%\\
1.6 m & 81.0\% & 100\% & 100\%\\
1.2 m & 46.3\% & 100\% & 98.6\%\\
0.76 m & 22.4\% & 47.6\% & 100\%\\
\midrule
0.76 m, localization noise (5 seeds) & 24.1\% & 46.0\% & 88.6\%\\
0.76 m, limits halved to 5 cm and $5^\circ$ & 9.5\% & 31.3\% & 100\%\\
0.76 m, reference executed on the robot & 46.9\% & 55.8\% & 100\%\\
\bottomrule
\end{tabular}
\caption{\rev{\textbf{Share of ledge crossings inside both entry limits.} 147 records per cell, 735 with noise.}}
\label{tab:distance}
\end{table}

\paragraph{\rev{Margin, noise, and cost.}} \rev{At 0.76\,m in kinematic simulation, the gate refuses every
approach outside the limits, the robot approaches again, and every rollout ends with a crossing
within them, none in a hold. Three further measurements put this result under stress
(Table~\ref{tab:distance}, lower rows). The margin is wide: the worst gated crossing
is $1.8\,\mathrm{cm}$ and $4.4^\circ$ off, and all 147 also meet limits half as large. With
Gaussian localization noise of $0.03\,\mathrm m$ and $1^\circ$ on the pose that the navigation
core sees (five seeds, 735 ledge records per arm), 88.6\% of records cross within the limits.
Of the 691 ledge approaches that are reached, 94.2\% cross within the limits, 5.5\% end in a
hold after two back-ups, and 0.3\% cross outside; without the gate, 54\% cross outside. On the
reference that completed the full route on the robot, the same three arms give 46.9\%,
55.8\%, and 100\%. The price is time: the mean time from offset to crossing rises from
$4.2\,\mathrm s$ to $11.3\,\mathrm s$.}

\paragraph{\rev{The gate as a test of the initiation set.}} \rev{The logs contain 1,470 ledge
approaches made with the approach window but without the gate, so nothing corrects the pose
after the check point; half of them ran under localization noise. Judged on the true pose at
the check point, all 350 approaches that satisfy Eq.~(\ref{eq:gate}) end within the limits; of
the 1,120 that fail it, 50.8\% do. The configured check is therefore a conservative inner estimate of the set of states from which the approach succeeds. Every approach it accepts in
these logs succeeds; the retries it adds are the time cost above. When the gate itself reads a noisy pose, 2 of 691 approaches cross outside the limits (0.3\%). A logistic boundary fitted to these approaches attributes the outcome
almost entirely to sideways offset and puts the limit at 27.5\,cm, close to the relaxed bound of 25\,cm (Appendix~\ref{app:isolation}).}

\paragraph{\rev{With learned skills in contact physics.}} \rev{In a controlled sweep of 378
contact simulations, our stairs policy passes 98.7\% of entries within $5^\circ$ of heading
error and 32.7\% of entries beyond $25^\circ$. Used as a classifier, the gate's limits separate
98.1\% passing inside from 72.2\% outside. Our learned-policy stack aligns to a stored direction before each stairs zone. In a matched
test with a sideways push 0.76\,m before each zone, the robot does not fall once in 180 climbs;
with only this alignment removed, 9 of 180 climbs fall ($p=0.004$). At the gabion ledge (0.32\,m), our ledge policy
falls in 5.3\% of 225 entries inside the gate's limits and in 12.3\% of 2,700 outside
($p=0.001$; Appendix~\ref{app:learnedskills}).}

\subsection{Rejoining the stored route}\label{sec:rejoin}
\leavevmode\rev{We keep the frozen centerline and live sensing fixed, and switch two factors on and off:
rejoining and the \emph{gait-entry lead}. With rejoining off, a robot that has been displaced
holds. The gait-entry lead is the factor that the appendices call entry conditioning: the
gait request is issued $0.6\,\mathrm m$ before the stored interval, and a rule that slows the
robot on entry is armed. Speed caps, gait labels, policies, and handoff logic are shared. Eight
segments, both factors, and matched displaced and undisplaced conditions are run with 23 seeds:
the three original seeds (192 runs) and 20 further seeds fixed before launch (1,280 runs). A $1.2\,\mathrm m$ sideways displacement is attempted at 8 m of
progress or 40\% of the segment length, whichever comes first, if the ground there is valid.}

\rev{The paired analysis needs every factor combination in a route--seed set to receive the
displacement. Six segments give 113 qualifying sets and 226 pairs with rejoining off and on.
We score a run by the competition's rule: the body must touch the 0.25 m disk of every goal of
its segment in order, and a fall voids the run. The benchmark's own rule is
stricter, the robot's centre within 0.20 m of each goal, a target 36\% smaller in area; we
report both. Completion rises from 0.4\% with rejoining off to 75.7\% with it on (interval 70.5
to 80.2 points), the level of the same sets without a displacement (77.4\%); under the stricter
rule the three figures are 0\%, 57.1\%, and 54.0\%. All 226 runs with rejoining return inside
the corridor and reach the next goal, against two of the 226 runs without it. Of the 55 that do
not complete, 54 fall at later transitions of the segment and one stalls, as undisplaced runs
do. A third arm replaces the hold with the tracker's default behaviour, a turn and a side-step.
On the 241 cells displaced in all three arms, completion is 0.4\% with the hold, 61.4\% with
that default, and 75.9\% with our strategy of walking forward while steering (difference 14.5
points, interval 8.2 to 21.3). After the push, our strategy reaches the next goal a median of
1.6\,s later than without a push; the default takes 5.8\,s longer. The gait-entry lead adds 6.2
points on undisplaced runs (interval $-0.5$ to 13.0; 8.2, 3.0 to 13.3, under the stricter rule). Displacements of 0.3 to 1.2\,m and runs with nothing injected
are reported in Appendix~\ref{app:recoveryextension}.}

\subsection{Route preparation and observation selection}\label{sec:routeprep}
\leavevmode\rev{Four configurations with one executor and the same passage fields compare how the centerline is prepared and what the planner observes. They are the stored route with its route-prior observations, the same route with live sensing, and two variants
that replace the centerline in the same corridor by shortest-path planning or a body-aware risk graph adapted from RB-TRG~\citep{guo2026rbtrg}. Over 2,800
valid trials, the live-sensing route completes 62.6\%, against 40.7\%, 44.0\%, and 43.6\% for
the other three, with paired gains over the two graph planners of 18.6 and 19.0 points
(Figure~\ref{fig:multiterrain-expanded}; Appendices~\ref{app:multiterrain},~\ref{app:geometry},~\ref{app:platform}, and~\ref{app:entryfactor}).}

\begin{figure}[t]
\centering\includegraphics[width=\linewidth]{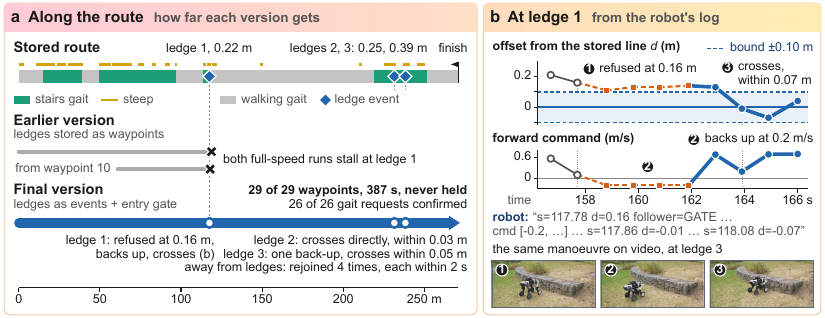}
\caption{\rev{\textbf{The robot's own log: earlier versions stop at the first ledge; the final version is refused
there, backs up, crosses, and finishes.} (a)~The stored route: gait intervals, steep ground, and three ledges.
An earlier version stored the ledges as waypoints; both of its full-speed runs stall at the first. The final
version stores each ledge as an event with an entry gate and reaches all 29 waypoints in 387 s without a
hold (Table~\ref{tab:robotlog}). (b)~The logged offset and forward command at the first ledge.}}
\label{fig:robotlog}
\end{figure}

\subsection{Deployment in a complete multi-obstacle competition environment}\label{sec:physical}
\leavevmode\rev{The organizers laid the route out across an open outdoor venue: 270.9 m in 28 segments,
with three ledges, sprint and turning sections, and more than ten kinds of terrain (stone slabs,
asphalt, terrazzo, grass, grass slopes, shrubs, a dry pond bed, a gabion, low and high stone
steps, mixed ground). People and props changed between runs, and a run counted only if the
robot reached every waypoint in order. On the competition day the final version, with LiDAR
localization and factory gait controllers (Figure~\ref{fig:physicaloverview};
Appendix~\ref{app:physical}), ran the full route six times and completed all six autonomously,
reaching all 29 waypoints in order and crossing every obstacle. It placed third in the
All-terrain Patrol task of the Embodied Future track at the 2026 Global Open-source AI
Challenge (GOAI; 2,999 valid preliminary submissions across all tracks;
\citealp{goai2026finalists}), and among that task's finalists it was the only entry to cover
the designated terrain sequence fully autonomously. The preliminary round ran the same navigation library in
simulation on a different map. The supplement has the reconstructed venue's simulation and one of the six runs on video.}
\rev{The robot's logs show how the entry gate works (Figure~\ref{fig:robotlog};
Appendix~\ref{app:robotlog}). In the logged end-to-end run of the evening before, the robot
reached 29 of 29 waypoints in 386.6 s; every velocity command came from the navigation node, and all 26 gait
requests were confirmed. At the first ledge it arrived 0.16 m off the stored crossing
line, was refused, backed up, and crossed within 0.07 m. It rejoined the route four times,
resumed within 2 s each time, and never held. On live-sensing
segments it detoured around unmapped obstacles, including team members moving near it (see the video).}

\section{Discussion}\label{sec:discussion}
\leavevmode\rev{A planner can avoid every obstacle and still deliver the robot to the next ledge a few degrees off, and a frozen skill library, factory or learned, does not say where each skill may begin.
One teaching drive does, once per transition: where and in which direction it was crossed. Enforcing it after every local replan keeps disturbed ledge crossings within the entry limits, stops our stairs policy from falling after a push, and lets displaced runs finish as well as undisplaced ones.}

\rev{It is also cheap by design: one teaching session, explicit rules, no training on site, and
the robot's own sensors and controllers. Wherever a robot must cross the same hard transitions again and again, one taught drive can hand the route
over to it: inspection rounds in plants and shipyards, patrols along mountain paths, and repeated supply runs through a surveyed disaster site.}

\rev{Three questions set our next steps. Can the gate's region be learned from the approaches every run logs, so that it refuses fewer good entries? Explicit rules decide which skill comes next and whether an obstacle can be stepped over; in clutter these become semantic questions, which we plan to distill into a small onboard model. Once a fleet shares stored routes, who takes which segment, and how do robots pass at narrow transitions? We release the stack and the benchmark so that such components can build on the same
interface. As skills ship with the robot or arrive as downloads, deployment may depend less on training new ones than on telling each robot where a skill may begin, a step toward robots that take over work that is dangerous for people.}

\clearpage
\section*{Reproducibility and AI use}\label{sec:statements}
The supplementary archive and the video are available at \url{https://drive.google.com/drive/folders/11vMdNe9M40steXCGX3HvS59p3p61cHaC}. The archive provides navigation and evaluation source code, count
tables, development and recovery run summaries, and the kinematic protocol and outcomes.
Failed outcomes and the setup-invalid ledger are retained. \rev{The supplementary material now
includes the two frozen policy files and the reconstructed scene, so the contact-simulation
benchmark can be re-run. The raw state trajectories of all runs and the robot logs are
omitted for size.}

Language models assisted manuscript organization and editing, literature and code
cross-checks, experimental diagnosis, and implementation of evaluation adapters,
instrumentation, and plotting scripts. We take responsibility for the research
claims and the resulting text, code, and analyses. Conceptual method and
terrain illustrations used an image-generation tool; empirical frames retain their original image content, with geometric annotations added as overlays.
Simulation figures render recorded states and measurements.

\bibliography{verified_refs}
\bibliographystyle{iclr2027_conference}
\clearpage
\appendix
\section{Separating the stored fields}\label{app:isolation}
{
\paragraph{Design.} Section~\ref{sec:isolate} uses the field navigation core with four
switches added. Each switch controls one of the mechanisms that the single ledge-handling
switch of the kinematic diagnostic (Appendix~\ref{app:kinematic}) turns on and off together. The first caps the approach speed. The second turns on
short-lookahead tracking over the stored approach window; tracking inside the platform-gait
interval stays on in every arm. The third is the entry gate of Eq.~(\ref{eq:gatepolicy}). The
fourth is the route-prior window after the ledge. With all four switches off, the core is
bit-identical to the handling-disabled arm of that diagnostic. With all four on, it is
bit-identical to the handling-enabled arm (294 crossing records, maximum difference 0).
Rollouts, offset injection, and scoring follow Appendix~\ref{app:kinematic}.

\begin{table}[!htbp]
\centering\small
\begin{tabular}{@{}ccccrrrr@{}}
\toprule
Speed cap & Tracking & Gate & Prior & Inside limits & Mean $|d|$ (cm) & Mean $|\Delta\psi|$ ($^\circ$) & Holds\\
\midrule
-- & -- & -- & -- & 22.4\% & 13.43 & 22.30 & 0\\
-- & -- & -- & on & 23.1\% & 13.42 & 22.21 & 0\\
-- & -- & on & -- & 93.9\% & 1.13 & 3.52 & 2\\
-- & -- & on & on & 93.2\% & 1.06 & 3.51 & 3\\
-- & on & -- & -- & 28.6\% & 9.76 & 21.64 & 0\\
-- & on & -- & on & 27.9\% & 9.65 & 21.58 & 0\\
-- & on & on & -- & 95.9\% & 0.87 & 2.70 & 2\\
-- & on & on & on & 95.9\% & 0.88 & 2.67 & 2\\
on & -- & -- & -- & 29.9\% & 8.82 & 20.00 & 0\\
on & -- & -- & on & 29.3\% & 8.82 & 20.02 & 0\\
on & -- & on & -- & 96.6\% & 1.28 & 4.37 & 0\\
on & -- & on & on & 96.6\% & 1.26 & 4.32 & 0\\
on & on & -- & -- & 47.6\% & 5.81 & 14.43 & 0\\
on & on & -- & on & 47.6\% & 5.79 & 14.40 & 0\\
on & on & on & -- & 100\% & 0.57 & 1.54 & 0\\
on & on & on & on & 100\% & 0.56 & 1.55 & 0\\
\bottomrule
\end{tabular}
\caption{\textbf{All 16 arms of the factorial.} Each arm has 49 offset cells and 147 ledge
records. The offset is injected 0.76 m before the ledge. ``Inside limits'' is the share of
records that cross within 10 cm and $10^\circ$; a ledge that is not crossed counts as outside.
``Holds'' is the number of the 49 rollouts that end in a hold. Errors are means over the
crossings that take place.}
\label{tab:factorial16}
\end{table}

\paragraph{Budget of back-ups and relaxed bound.} Table~\ref{tab:gatebudget} varies the two
exception settings of the gate, the number of back-ups allowed and the relaxed bound, with
all other mechanisms on (441 rollouts). Without a back-up, the gate can still pivot the
robot to fix its heading, but it cannot fix a lateral offset. The robot then holds, or a wide
relaxed bound lets it cross outside the limits. With at least one back-up, no crossing fell
outside the limits in any setting. A robot that the relaxed bound lets through is at most
$\delta_r$ off the line at the check point, and tight tracking over the last 0.7 m removes the
rest.

\begin{table}[!htbp]
\centering\small
\begin{tabular}{@{}lrrr@{}}
\toprule
Back-ups allowed $R$ & $\delta_r=0.15$ m & $\delta_r=0.25$ m & $\delta_r=0.35$ m\\
\midrule
0 & 34.7\% (30 holds) & 38.1\% (27 holds) & 49.7\% (16 holds; 29 outside)\\
1 & 53.7\% (23 holds) & 87.8\% (7 holds) & 100\% (0 holds)\\
2 & 91.8\% (5 holds) & 100\% (0 holds), configured & 100\% (0 holds)\\
\bottomrule
\end{tabular}
\caption{\textbf{Gate exceptions.} Each cell gives the share of the 147 ledge records inside
both limits and the number of the 49 rollouts that end in a hold. ``Outside'' counts crossings
outside the limits. It is zero in every cell with $R\ge1$.}
\label{tab:gatebudget}
\end{table}

\paragraph{Localization noise.} At every control tick we add Gaussian noise to the pose that
the navigation core receives: standard deviations of 0.03 m in planar position and $1^\circ$
in yaw, the pose-noise condition of the contact simulation. Scoring uses the true pose. Five
seeds give 245 rollouts and 735 ledge records per arm. Conventional tracking brings 24.1\% of
records inside the limits, the approach window 46.0\%, and the gate 88.6\% (per-seed range
87.1--89.8\%). The paired difference between the last two is 42.6 points (seed bootstrap
41.6--43.7). With the gate, the robot reaches 691 of the 735 ledge approaches. The other 44
are not reached because the rollout already ended in a hold at an earlier ledge. Of the 691,
651 cross inside the limits (94.2\%), 2 cross outside (0.3\%), and 38 end in a hold after two
back-ups (5.5\%). All 38 holds are at a lateral offset of 0.3 or 0.4 m. The offset trigger
reads the noisy route progress. It misses its 8 cm window in one of the 691 gated approaches
and in 23 conventional-tracking records, and we score these records as they ran. With the
limits halved to 5 cm and $5^\circ$, the three arms give 12.0\%, 33.7\%, and 83.1\% with
noise, and 9.5\%, 31.3\%, and 100\% without.

\paragraph{The gate check as a classifier.} We took every ledge approach of the conventional
and approach-window arms (2,940 across the factorial, the distance sweep, and the noise runs).
For each, we read from the tick logs the true pose at the check point and the outcome at the
ledge. With the approach window on (1,470 approaches, 735 of them from the noise runs), 350
approaches pass the check of Eq.~(\ref{eq:gate}), and all 350 end inside the limits: 274 of
274 without noise and 76 of 76 with noise. Of the 1,120 that fail it, 569 (50.8\%) end inside.
Under conventional tracking the same figures are 214 of 246 (87.0\%) and 357 of 1,224
(29.2\%). The check predicts the outcome under conventional tracking as well. With the
approach window, passing the check is enough to cross inside the limits; under conventional
tracking the pose still drifts over the last 0.7 m. We also fitted a logistic boundary on
lateral offset and heading error at the check point to the approach-window data. Its
five-fold accuracy is 76.6\%, against 61.3\% for the configured check used as a classifier.
Its weights are $-0.85$ per 10 cm and $-0.05$ per $10^\circ$, and it crosses zero at 27.5 cm
when the heading error is zero. Heading error at the check point carries almost no weight,
because the robot corrects it over the remaining distance.

\paragraph{The reference executed on the robot.} The reference evaluated above derives from
the one that completed the full route on the robot and has the same three ledge crossings. On
the robot's own reference, the three main arms (conventional tracking, approach window, entry gate) give 46.9\%, 55.8\%, and 100\% (mean errors
5.78 cm and $15.21^\circ$; 4.23 cm and $11.14^\circ$; 0.70 cm and $2.01^\circ$; no hold).
}

\begin{figure}[!htbp]
\centering\includegraphics[width=\linewidth]{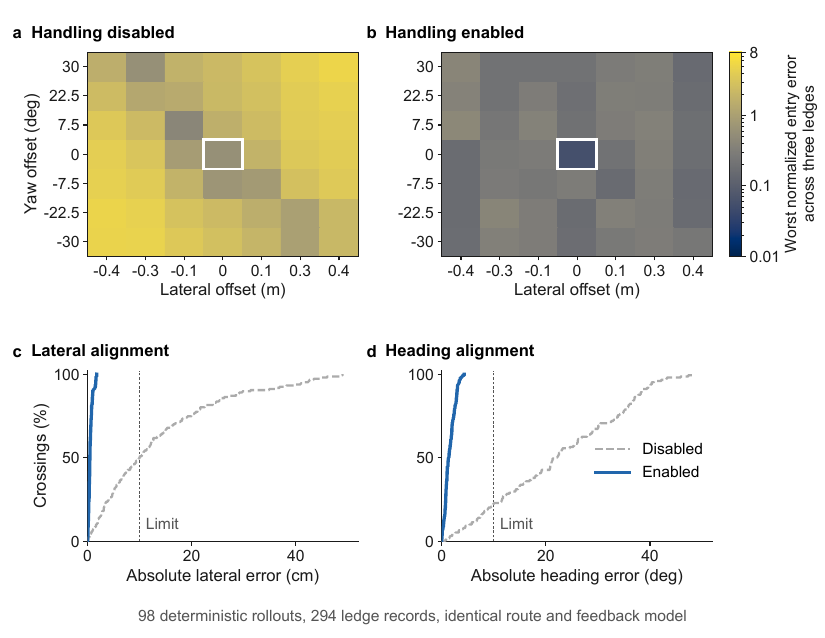}
\caption{\rev{\textbf{Crossing errors at the three ledges.} Top: for each offset cell, the
worst normalized error among the three ledges, on a shared logarithmic colour scale.
For each crossing, $E=\max(|d|/(0.10\,\mathrm m),|\Delta\psi|/(10^\circ))$;
$E\le1$ means that both configured limits are met. White boxes mark zero-offset controls.
Bottom: cumulative distributions of the absolute crossing errors over all records.
The kinematic model delays the commands and feeds back a synthetic gait.}}
\label{fig:ledgeconditions}
\end{figure}

\section{Preparation effort and pre-deployment checks}\label{app:preparation}
{
Teaching takes one session. The operator drives the route two or three times, about six
minutes per pass (three passes here), and autonomous runs on the compiled reference began the
same day. The passes fix the corridor that the compiler may use, and the last two supply the
gait intervals. The reference takes its line from the recorded passes, so teaching needs one
clean traversal of each passage. Where a passage needs a better line, the operator drives that
passage again and the compiler splices the new recording in (five times here). Recording a re-drive takes as long as driving the passage; the five cover 16 to 111 m. In all, the recordings that the reference uses take 27.2 minutes of driving. Compiling involves
no training and no parameter fitting. Each ledge is crossed in one of the spliced re-drives, so
its crossing position and direction come from one recorded traversal. The entry limits, speed
caps, and retry count are configured values shared by the whole environment; lane lengths are
set per ledge. A
new route needs new demonstrations and a recompile. A terrain change at an existing ledge needs
a new recording of that passage. Corridor widths and tolerances are configuration edits.

The contact-simulation stack and the field stack share the same navigation library byte for
byte (35 source files). In the simulation runs, 106 of the 116 runtime values are identical.
The others are speed settings, the body-height offset, the control rate, two ledge-lane values
that the simulation reference does not use, and the terrain-prior switch varied in
Section~\ref{sec:routeprep}. The top-level core file of the simulation snapshot differs from
the field one in 37 lines, none of them on the planning path. The simulation snapshot predates
the route-prior window after a ledge, the gait-interval hysteresis, and a pose logger.

Before installation, every candidate reference goes through two checks. The first sweeps a
$0.9\times0.5$ m body footprint along the route and checks its clearance against the map. The
second is a kinematic run of the full route, which must reach all waypoints. Over the two days
of preparation, the kinematic run failed 13 times and the clearance check reported violations
11 times. Every installed revision passed both.

\begin{table}[!htbp]
\centering\small
\begin{tabularx}{\linewidth}{@{}p{0.34\linewidth}>{\raggedright\arraybackslash}p{0.3\linewidth}>{\raggedright\arraybackslash}X@{}}
\toprule
Item & Source & Quantity for the 270.9 m field route\\
\midrule
Centerline, waypoint entries & automatic, from pose trails and marks & 3 full demonstrations (6.1, 6.1, 5.9 min); 29 waypoint entries; 28 segments\\
Local corrections spliced into the line & operator selects recordings; splicing is automatic & 11 re-drives recorded, 5 selected (26, 32, 111, 57, 16 m; 1.5, 0.8, 4.6, 1.5, 0.7 min)\\
Ledge crossing position and direction & automatic, from platform-gait trail events with optional map adjustment & 3 ledges\\
Gait intervals & automatic, from recorded gait events and map steepness & 9 intervals (6 stairs, 3 platform)\\
Compiler settings & operator, one set for the environment & 31 scalar values\\
Runtime settings & shared configuration & 116 values\\
\bottomrule
\end{tabularx}
\caption{\textbf{What a reference costs to prepare.} Durations are measured from the
recorded pose trails, for the recordings that the reference uses. The one-time map recording
took 25 minutes.}
\label{tab:preparation}
\end{table}
}

\rev{\paragraph{Planning the route from the map alone.} Before we built the teaching compiler,
we tried to plan the route for the same venue from the recorded point-cloud map and the official waypoint sequence of that stage (30 waypoints and 29 segments). We wrote two grid
planners for this. One searches over the flat cells of each segment, and the other searches
jointly over cells and gaits. With its supplied settings, the joint planner returned a
candidate for 26 of the 29 segments. It did so only because obstacle cells were cleared
wherever an earlier trajectory had passed. Without that override it returned 13 of 29, and with
a body clearance of 0.4 m added, 4 of 29. In 20 of those 25 failures, the waypoint cell itself
was not passable in the map. With one shared conservative traversability mask, neither planner
connected any of the eight flat segments where re-routing is allowed, each time because a
waypoint cell was blocked. No setting gave a usable route across all 29 segments. The map
covers the venue unevenly: 51\% of its 0.1 m cells are known, and 31\% are directly observed.
A taught traversal does not depend on map quality at any waypoint, because the driven line
itself shows that the ground is passable. Changes also stay local. When a waypoint or a passage
changes, the operator re-drives that passage and the compiler splices it in, without recording
the map or the rest of the route again; this was done five times for the evaluated reference. The organizers revised the waypoint
sequence several times during preparation, and each revision was absorbed in this way, with nothing
retrained and no gait retuned.
The map-based route had to be planned again from the start. A fixed route also failed in
execution on the routes then in use: in contact simulation, route following without recovery
completed the full route in 4 of 24 seeds (Appendix~\ref{app:learnedskills}). The task leaves
little margin. Every waypoint is a 0.25 m disk that must be reached in order over 270 m, and in
our experience a fixed route that works in a small scene does not hold up at this scale. The
taught reference adapts during the run through detours, the entry gate, and rejoining. The
result files are in the supplementary material (\texttt{results/map\_only\_planning}).}

\section{Component removals and execution sensitivity}\label{app:component}
\rev{We connected the navigation core to existing MuJoCo scenes and to the frozen actors
\texttt{J3100.onnx} and \texttt{stairs\_1150.onnx}. The simulation adapter is separate from the
factory locomotion stack used on the physical robot. We test three route segments under
nominal conditions, pose perturbation, and an inserted obstacle, with seeds 11, 23, and 37.
Every segment--condition--seed cell runs all five configurations, for 135 attempts. These are
development experiments on the existing reconstructed scene.}

\rev{The five configurations are the current core and four variants. Each variant removes one
component: local detours, entry braking, route readjustment, or the adapter's
terrain-observation override. Readjustment is the rejoining of Section~\ref{sec:rejoin}, and
the terrain-observation override supplies the route-prior observations. Without detours, the
robot may not detour sideways and has no A* fallback, but it keeps live observations and
readjustment. Disabling readjustment changes the route-rejoining procedure. Without the terrain
override, this adapter keeps the live-input path. Each variant tests one execution interface,
and the other reference fields stay fixed.}

The file \texttt{component\_trial\_records.csv} in the source data holds the complete
trial-level development outcomes.

\begin{table}[!htb]
\centering\small
\begin{tabular}{@{}lrrrr@{}}
\toprule
Removed component & Completion & Falls & Obstacle contact & Non-wheel contact\\
\midrule
local detour & -11.1 & +0.0 & +7.4 & +7.4\\
entry braking & +7.4 & -11.1 & +0.0 & -3.7\\
readjustment & -3.7 & +3.7 & -14.8 & -11.1\\
terrain override & +18.5 & -11.1 & +0.0 & -14.8\\
\bottomrule
\end{tabular}
\caption{\rev{\textbf{Effect of removing each component in the development batch.} Entries
are percentage-point changes from the current core on matched cells, with equal weights for
the three routes and the three conditions. A positive completion change means more completed
tasks; a positive event change means more recorded events. The batch has 135 attempts.
Contacts and falls are sampled at $50\,\mathrm{Hz}$ and are not mutually exclusive. The full
counts are in \texttt{component\_outcome\_counts.csv}.}}
\label{tab:development}
\end{table}

\rev{Local detours raise aggregate completion by 11.1 percentage points over disabling them.
Removing entry braking or the terrain override can raise completion. These effects depend on
how the execution rules interact with the fixed actors. The paired table keeps every tested removal. Removing readjustment lowers obstacle-contact
incidence by 14.8 points and completion by 3.7 points. Where both configurations succeed, the
paired completion-time differences have a median of zero. Falls, blocked states, and timeouts
are in the full source counts.}

\paragraph{Outcome accounting.}
\rev{Ordered completion uses the frozen goal coordinates and their order. We report contact,
fall, and termination events for every attempt. We compare completion times only within
matched cells where both configurations complete, and we report that denominator. Controller,
scene, task, and perturbation settings stay fixed within a comparison. The component-removal
batch is kept apart from the structural comparisons and the displacement diagnostic, and its
contact and termination statistics are reported separately.}

\subsection{Execution-environment and seed comparisons}\label{app:platform}
\begin{table}[!htbp]
\centering\small
\begin{tabular}{@{}lrrrr@{}}
\toprule
 & Prior reference & \multicolumn{3}{c}{Change from prior reference (pp)}\\
Batch & Completion (\%) & Live sensing & Shortest path & Body-aware\\
\midrule
macOS & 70.4 & -7.4 & -14.8 & -18.5\\
Linux recheck & 51.9 & +18.5 & +11.1 & +14.8\\
Linux new seeds & 48.1 & +11.1 & +14.8 & +18.5\\
\bottomrule
\end{tabular}
\caption{\rev{\textbf{Completion per batch and paired differences between configurations.}
The three batches contain 324 attempts in total. Within each batch, three route segments and
three conditions have equal weight, with three seeds per cell and four configurations. Each
difference subtracts the prior-enabled reference of the same batch. We keep the batches
separate because their execution environments and seed sets differ.}}
\label{tab:baselines}
\end{table}

\rev{The structural study has 108 macOS attempts, 108 Linux attempts that recheck the exact
same sources, and 108 further Linux attempts with new seeds. The recheck uses the same pinned
source and asset files as the macOS evaluation. macOS arm64 runs Python 3.12.14, and Linux
x86\_64 runs Python 3.10.16. MuJoCo 3.3.1, ONNX Runtime 1.19.2, NumPy 1.26.4, and SciPy 1.13.1
are the same on both. The packaged sources are held fixed, while the operating system,
processor architecture, interpreter, and numerical effects change together. The new-seed batch
uses the earlier baseline runner on the same reconstructed scene and is reported separately.}

\rev{The source-data file \texttt{development\_condition\_counts.csv} holds the condition-level
counts. In each batch, the graph adapter completes more inserted-obstacle cases than the
live-sensing reference. The batch-level differences in Table~\ref{tab:baselines} show the
reversal across execution environments.}

\subsection{Entry timing and observation selection}\label{app:entryfactor}
\rev{On segment WP05--WP08 we cross three factors, terrain-prior selection, the gait-entry
lead, and entry deceleration, under nominal execution and pose noise, with seeds 11, 23, and
37. The full design has 48 attempts: 30 further runs and 18 matching records from the
component study. Scene, task, route, and packaged sources are the same, and the navigation
settings differ only in the three factors. The lead issues the gait request 0.6 m early.
Deceleration is either off or set to 0.8 in the existing entry rule. All other execution
settings are shared.}

\begin{table}[!htbp]
\centering\small
\begin{tabular}{@{}lr@{}}
\toprule
Observation & Mean completion effect of a 0.6 m entry lead (pp)\\
\midrule
Live sensing & +25.0\\
Route prior & $-50.0$\\
\bottomrule
\end{tabular}
\caption{\rev{\textbf{The effect of entry timing depends on the observation source.} Each
effect subtracts the no-lead setting on matched cells. The nominal and pose-noise conditions,
and deceleration off and on, have equal weights. The design has 48 executions, 18 of them
reused records. The effects are for this fixed-segment diagnostic. All sixteen condition-level
counts are in \texttt{entry\_factor\_counts.csv}.}}
\end{table}

\section{Implementation and controller interfaces}\label{app:implementation}
\rev{\paragraph{Details of compilation and execution.} Candidate ledge crossings come from the
parts of the teaching trails driven in the platform gait. The local planar displacement there
gives the approach direction. Clustered detections give an estimate of the crossing position,
which the compiler can adjust with the ground-cloud profile. The compiler blends a configured
straight crossing lane into the route only when the shift, waypoint-access, and body-sweep
checks allow it. Gait votes from the demonstrations are projected onto the assembled line,
combined with map-derived steepness and configured platform extents, and turned into intervals
along the arc length. Acceptance checks can reject a candidate reference. Hazard and skew
diagnostics record further geometric conditions, and the field build runs a body-sweep
clearance check on each reference before its kinematic route check. In the field reference, the
crossing lane before the ledge, $b_j$, is 2.0\,m at two ledges and 0.7\,m at the first, and
$a_j$ is 1.0\,m. At run time, $S_t$ holds a bounded history of local observations, route
progress, tracking error, the selected lateral-transition state, and the handoff and blocked
timers. The observation history is cleared by age or at discontinuities. The transition anchor
of the retained candidate persists between ticks, so a lateral motion is not restarted. The
fixed simulation actors also take the previous action and gait phase as input
(Appendix~\ref{app:actorinterface}). Gait-interval hysteresis keeps the current interval when
projected progress moves slightly backward.}
\begin{table}[!htbp]
\centering\small
\begin{tabularx}{\linewidth}{@{}>{\raggedright\arraybackslash}p{0.20\linewidth}>{\raggedright\arraybackslash}X>{\raggedright\arraybackslash}X@{}}
\toprule
Information & Origin & Runtime use\\
\midrule
Centerline and goals & Selected pose trails, passing marks, map-checked geometric processing & Route projection and candidate coordinates\\
Corridor and detour permission & Configured widths and permissions, with clearance checks & Bounds on local alternatives\\
Ledge event and approach pose & Platform-gait trail events, local displacement, optional map adjustment & Event-specific approach tracking and checks\\
Gait intervals & Recorded gait votes, map steepness, configured platform extents & Requested locomotion mode and boundary timing\\
Speed caps and approach extents & Configured route fields and runtime settings & Command limits and activation windows\\
Retry and handoff rules & Shared execution configuration & Re-approach, hold, and switch coordination\\
\bottomrule
\end{tabularx}
\caption{\rev{\textbf{Information supplied to the executor.} Fields derived from the
demonstrations are kept apart from configured rules. A route bundle stores the information of each event, and shared runtime settings decide how the executor uses it.}}
\label{tab:fieldorigins}
\end{table}
\rev{The field implementation compiles pose trails, waypoint marks, and gait events into a
route and terrain bundle that can be reloaded. Corridor widths and speed caps are configured.
Crossing geometry and gait intervals combine recorded events with geometric rules. Route
progress keeps the projection from jumping to a nearby return leg. A bounded observation queue,
the retained lateral-transition anchor, and gait-boundary hysteresis keep the state continuous
within a run.}

\rev{On the robot, the executor can request the flat, stairs, and platform gaits. The simulation
comparison uses the flat and stairs actors, frozen as \texttt{J3100.onnx} and
\texttt{stairs\_1150.onnx}. The evaluation manifests list the actor hashes and environment
versions. Appendix~\ref{app:archivedtraining} documents the archived controller-training
configuration.}

\rev{The graph baselines sample the same frozen simulation corridor every 0.5 m, with five lateral
offsets at fractions $-1,-0.5,0,0.5,1$ of its half-width. Nodes in adjacent layers are joined
only where the sampled map cells are known and free of static obstacles. The shortest-path
variant minimizes edge length. The RB-TRG adapter runs its body-aware planner on this graph and
the common elevation grid, with a $1.0\times0.6$ m body footprint. Only the centerline
coordinates are replaced; segment endpoints and the other execution fields stay as they are.
The adapter places the repository's planning core inside our shared executor.}

\rev{For Figure~\ref{fig:rejoin}, we project each recorded robot position onto every nonzero-length
line segment of that run's frozen original centerline and plot the minimum point-to-segment
distance. The distance is therefore always measured to the same line, even while rejoining changes
the path coordinates that the follower tracks.}

\section{Instrumented entry-conditioning diagnostic}\label{app:entryforce}
\rev{This diagnostic tests whether the entry-conditioning branch (the gait-entry lead of Section~\ref{sec:rejoin}) intervenes during runs or is only configured. We add logging to the
branch without changing any command. On the first route, seeds 11, 23, and 37 cross entry
conditioning on/off with a fixed force on/off, for 12 cells. The reference, the live-sensing
permissions, the actors, the goals, and the enabled rejoining procedure stay fixed. At 8 m of
route progress, a force of $0.1mg=18.63$ N pushes the robot to the left for 0.3 s; the model's
total mass is 18.99 kg. All six force-on cells record the force event, and all 12 cells
complete.}

\rev{The logs record the command before and after conditioning, the distance to the next
boundary, the computed speed threshold, and a flag that shows whether braking replaced the
command. With conditioning on, the threshold is computed on 719 ticks without the force and
725 ticks with it, summed over the three seeds of each group. No command exceeds its threshold,
so braking replaces no command in any cell. The gait lead still changes the time of the first
transition. The combined setting therefore has two parts: the early gait request acts, and the
deceleration rule never fires on this route, with these actors, at these speeds.}

\section{Teaching traces and local compilation}
\begin{figure}[!htbp]
\centering\includegraphics[width=\linewidth]{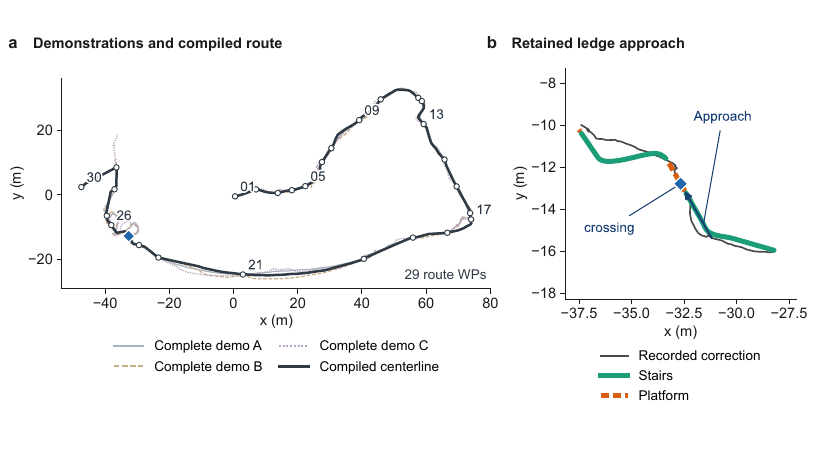}
\caption{\rev{\textbf{From recorded passes to the stored route.} (a) The three complete
teaching traces and the compiled route with its 29 waypoint entries. (b) A recorded local
correction near a ledge, with the compiled crossing, approach direction, and gait intervals.
The curve and the fields are the compiler's stored output.}}
\label{fig:teaching}
\end{figure}
\section{Kinematic diagnostic on the compiled field reference}\label{app:kinematic}
\paragraph{Geometry and event information.}
\rev{The compiled centerline has 2,711 XY samples and is 270.923 m long. Each of the three
stored crossing positions lies within 0.4 mm of this polyline. At each marked crossing, the
tangent over a symmetric $\pm0.25$ m arc window agrees with the stored heading to within
$0.01^\circ$; over a $\pm1$ m window it differs by at most $1.48^\circ$. These values show
that the saved curve and the labelled events agree with each other. The event records give the
ledge locations and approach windows, and configured rules set the checks, speed limits, and
retries (Table~\ref{tab:fieldorigins}).}

\paragraph{Rollouts and measurements.}
\rev{The diagnostic loads the field route, the terrain records, the gait plan, and the
navigation core, and visits the three stored ledges in route order. The main range-extension
study crosses ledge handling on/off with the seven lateral and seven heading offsets given in
the main text. This gives 98 deterministic rollouts and 294 ledge records. An initial
diagnostic used lateral offsets $0,\pm0.20$ m and heading offsets $0,\pm15^\circ$, which gave
18 rollouts and 54 records. In it, 37.0\% of records met the geometry limits with handling
disabled and 100\% with handling enabled (27 records per setting). We report the two grids
separately. Each offset is applied once per ledge, when projected progress first lies
0.72--0.80 m before the ledge's stored arc coordinate. In the main grid, six records are
zero-offset controls and 288 receive a nonzero offset. We keep all records; the three ledge
measurements of a rollout are grouped together.}

\rev{The planar pose is integrated from body-frame velocity commands at 20 Hz. After seven
initial ticks with zero command, commands take effect with a steady-state delay of 0.30 s. The
pose input of the core reports zero velocity. The gait starts as flat, and a local feedback
model switches the reported gait once a request has been pending for more than 1.5 s. In both
settings, the flat and stairs speed overrides and the walking speed are 1.0 m/s; the
near-ledge limits stay active when ledge handling is enabled. The diagnostic tests the
runtime decisions with kinematic integration and synthetic command and gait feedback.}

\rev{Both settings receive the same flat-observation input and the same gait-zone prior
configuration. The enabled setting also uses the route-prior window after the ledge. Its factor
therefore combines the stored crossing geometry, approach tracking, speed limits, retry
behaviour, and prior selection. These mechanisms change together, and prior selection can
change the observations received during execution. Appendix~\ref{app:isolation} separates
them.}

\rev{For the next ledge in order, the crossing is the first sample that meets three
conditions: projected progress within 0.35 m of the ledge's stored arc coordinate, a forward
coordinate in the ledge frame in $[0,0.30)$ m, and an absolute lateral coordinate below 1.5 m.
This rule excludes unrelated intersections of the route. We then test the recorded pose
against the configured limits of $0.10$ m lateral and $10^\circ$ heading. All 294 main-grid
records have a crossing, and all offsets were applied. A rollout ends on completion, on a hold,
or when progress passes 260 m. The diagnostic scores geometry at the three ledges before the
260 m cutoff. We logged poses, commands, gait requests, synthetic feedback, and crossing events
at every tick.}

\clearpage
\section{Physical terrain coverage and approach sequence}\label{app:physical}
\begin{figure}[!htbp]
\centering\includegraphics[width=\linewidth]{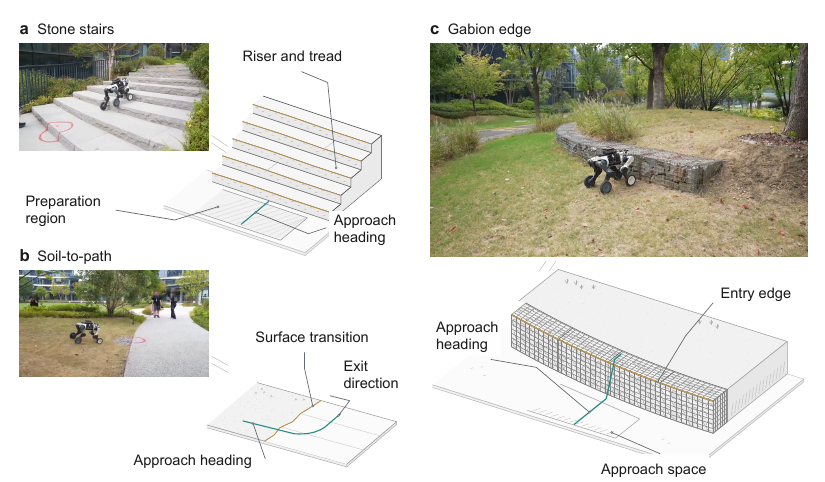}
\caption{\rev{\textbf{Field scenes and their passage geometry.} (a) Stone stairs, (b) a
soil-to-path transition, and (c) a gabion ledge, each shown with a sketch of the passage
geometry that matters there. Amber marks visible terrain boundaries. Teal arrows show approach
directions on schematic diagrams that are not to scale. Bystanders' faces are pixelated.}}
\label{fig:physical}
\end{figure}

\begin{figure}[t]
\centering\includegraphics[width=\linewidth]{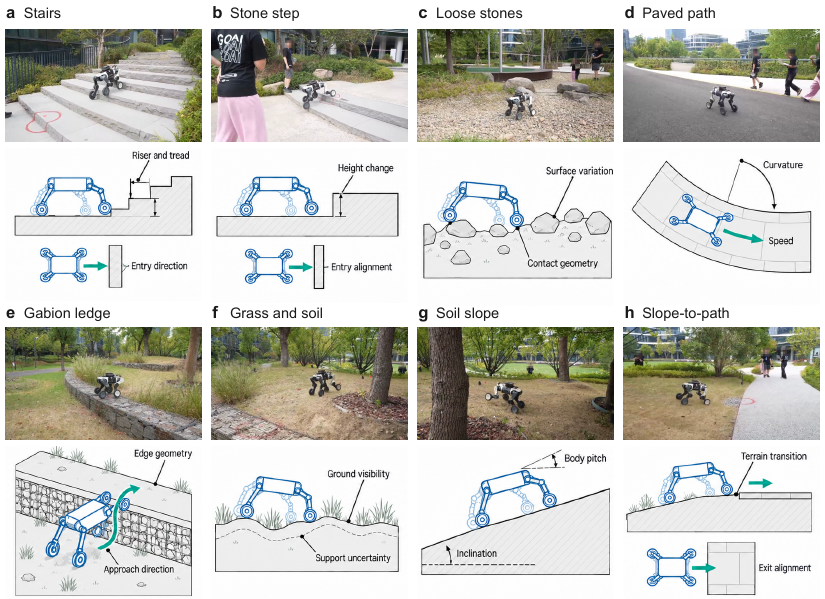}
\caption{\rev{\textbf{Terrain transitions and their passage conditions.} Each pair shows a
field scene above and its geometry below: height and entry direction, local surface variation,
route curvature, and slope transitions. The schematics are not to scale. Bystanders' faces are
pixelated.}}
\label{fig:physicalpairs}
\end{figure}

\begin{figure}[!htbp]
\centering\includegraphics[width=\linewidth]{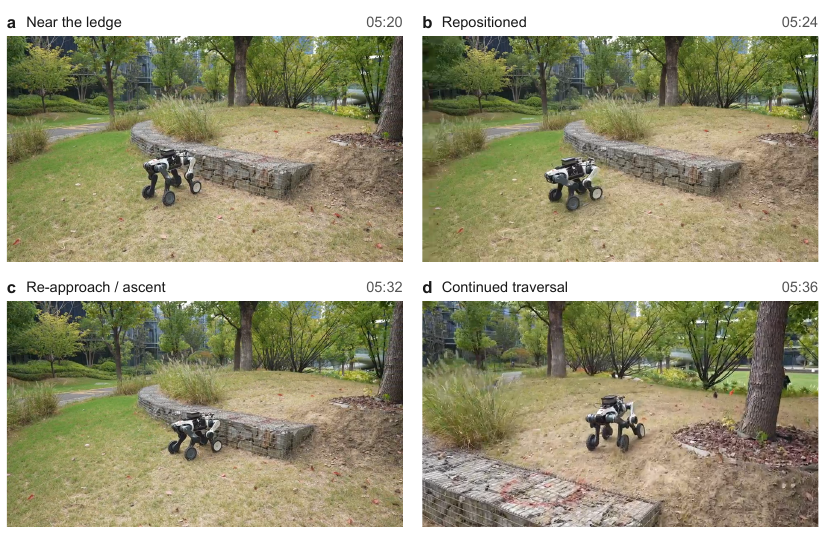}
\caption{\rev{\textbf{Ledge approach and repositioning.} Video frames of the robot show the
approach, the repositioning, the second approach and ascent, and the traversal that
follows.}}
\label{fig:sequence}
\end{figure}
\section{\rev{Robot log of the end-to-end run}}\label{app:robotlog}
\rev{The navigation node logs its state about once per second: commanded velocity, distance
along the reference, lateral offset from the reference, planner state, requested gait, and
every state change. The control node logs each gait request, each confirmation by the factory
controller, and the source of each velocity stream. Offsets come from the on-board LiDAR
localization. The log has no heading channel, so we report only lateral offsets.

\paragraph{The gate on the robot.} In the end-to-end run (29 of 29 waypoints, 386.6 s,
0.70 m/s on average) the robot handled the three ledges as follows. At the first ledge (rise
0.22 m), the first approach reached the check 0.16 m off the stored crossing line, against a
bound of 0.10 m. The gate refused it. The robot backed up at 0.2 m/s over the 0.5 m configured
that evening (0.3 m in the final configuration), came back 0.01 m off the line, and crossed
with at most 0.07 m of offset. It crossed the second ledge (0.25 m) on the first approach,
within 0.03 m. At the third ledge (0.39 m), the robot pivoted in place at the check, drifted to
0.15 m, backed up once, and crossed within 0.05 m. Figure~\ref{fig:sequence} shows such a
sequence in video frames. Away from the ledges, the robot entered the re-adjust state, in which
it rejoins the reference, four times: once for a blocked centerline, once for a planned path
that failed the footprint check, and twice after 8 s without progress. Each time it resumed
tracking within 2 s. It never entered the hold state before the finish. All 22 velocity
streams came from the navigation node, the operating mode did not change during the run, and
all 26 gait requests were confirmed. The daytime runs of the same day predate the re-adjust
state and show what it replaces. Nine of the 13 logged runs on the route ended in a hold: four
because the robot was more than 0.8 m off the route, and five after 8 s without progress. The
end-to-end run met the no-progress condition twice and cleared it through the re-adjust state.
This state is the on-robot counterpart of the rejoining factor in Section~\ref{sec:rejoin}.

\paragraph{From the development versions to the final one.} Table~\ref{tab:robotlog} lists, in
order, every logged run of that day that approached the first ledge. Rows 1 to 3 are earlier
development versions. Rows 4 to 6 are the version with ledge events, which we finished that
night and used on the competition day. With the ledge stored as an ordinary waypoint, the
robot crossed it only when it drove the whole route slowly. At the full speed setting, it
reached the waypoint, turned toward the next one while still against the ledge, and stalled.
With the ledge stored as an event, the first run was already aligned within 0.05 m. It did not
mount the ledge at a crossing speed of 0.4 m/s, so we raised the cap to 0.7 m/s, and the
following runs crossed. The walking speed is a platform setting. That day we raised it from 1.0
to 1.67 m/s in steps, because the platform's sensor timestamps occasionally fell out of
alignment at high speed. Nothing was retrained. Runs 1 to 3 predate the re-adjust state and ran
without the obstacle map. The table records that day in development order; the matched
comparison is the factorial of Section~\ref{sec:isolate}.

\paragraph{Competition day.} On the competition day, the final version ran the full route six
times and completed all six autonomously, each time reaching all 29 waypoints in order. Under
the competition's scoring, a run is void if it misses any waypoint. These runs are recorded on
video. The log figures above come from the evening before. The competition's preliminary round
was held in simulation, in an environment built by the organizers whose map differs from the venue
as built and from the reconstructed scene used in Section~\ref{sec:rejoin}. The same navigation library
qualified for the final there, so the organizers scored it in two independently prepared
environments, one in simulation and one on the robot.}

\begin{table}[!htbp]
\centering\small
\caption{\rev{\textbf{Logged runs that approached the first ledge on the day before the
final.} Rows 1 to 3 are earlier development versions, and rows 4 to 6 are the final version. Times are local. Speed is the walking-speed setting. Offsets are lateral distances
from the reference.}}
\label{tab:robotlog}
\rev{\begin{tabular}{@{}ll>{\raggedright\arraybackslash}p{0.8cm}>{\raggedright\arraybackslash}p{5.0cm}>{\raggedright\arraybackslash}p{3.5cm}@{}}
\toprule
Time & Ledge stored as & Speed (m/s) & At the first ledge & Run outcome \\
\midrule
10:37 & waypoint & 1.0 & crossed at 0.3 m/s in the stairs gait & finished from waypoint 10 (215.6 m) in 572 s, 0.38 m/s \\
16:53 & waypoint & 1.67 & stalled for 11 s, turn command $-0.8$ to $-1.0$ rad/s, 0.09 to 0.30 m off the line & paused by the operator \\
18:33 & waypoint & 1.67 & stalled, turn command at its limit, ended 0.35 m off the line & held \\
23:18 & event & 1.67 & aligned within 0.05 m; did not mount at 0.4 m/s & held after 20 s without progress \\
23:36 & event & 1.67 & crossed & log file ends at 160 m \\
23:52 & event & 1.67 & refused at 0.16 m, one back-up, crossed within 0.07 m & 29 of 29 waypoints in 386.6 s, no hold \\
\bottomrule
\end{tabular}}
\end{table}

\clearpage
\section{Map-ground sections and spatial method illustration}
\begin{figure}[!htbp]
\centering\includegraphics[width=0.94\linewidth]{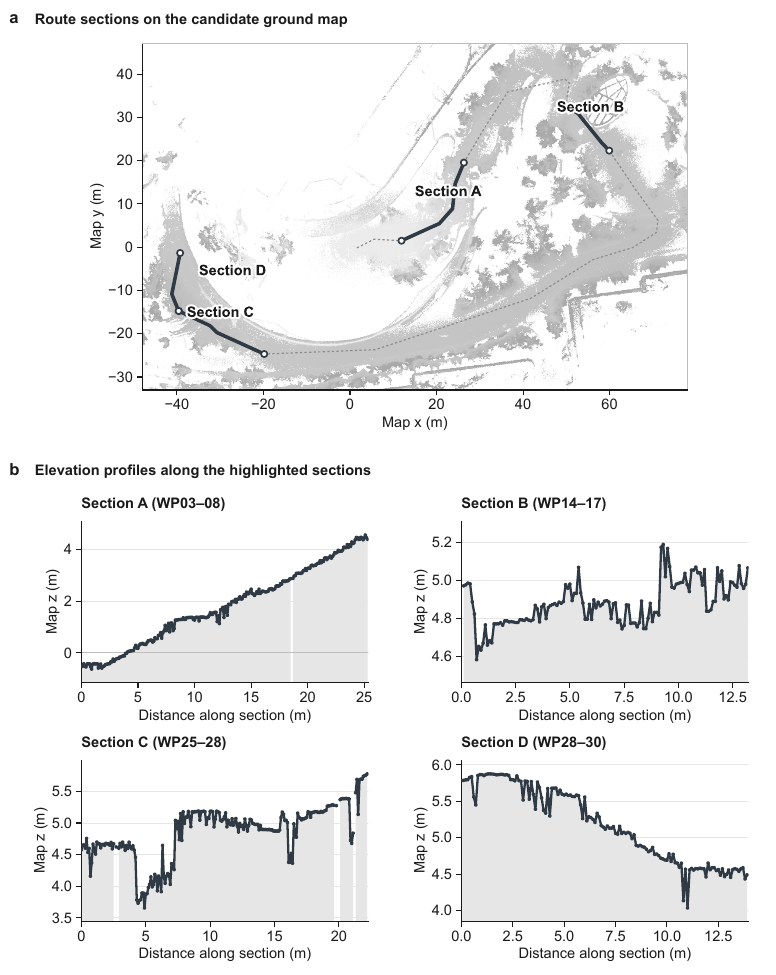}
\caption{\rev{\textbf{Terrain along the route and local elevation profiles.} A--D mark four
sections of the reference route on a grey map, in the same colours as their profiles. The
0.1 m import surface takes the per-cell minimum of the archived point cloud, over cells with at
least two returns. Each section samples the frozen reference XY every 0.1 m and uses only known
cells. The sections have 254, 133, 223, and 140 stations; missing map samples are left as gaps.
Panel scales differ. The profiles combine the XY coordinates of the simulation reference with
elevations from the candidate ground map. Sparse scans, vegetation, and multiple surfaces can
affect the cell minima. The field video is not registered to these section coordinates.}}
\label{fig:terrainsections}
\end{figure}
\clearpage
\begin{figure}[!htbp]
\centering\includegraphics[width=\linewidth]{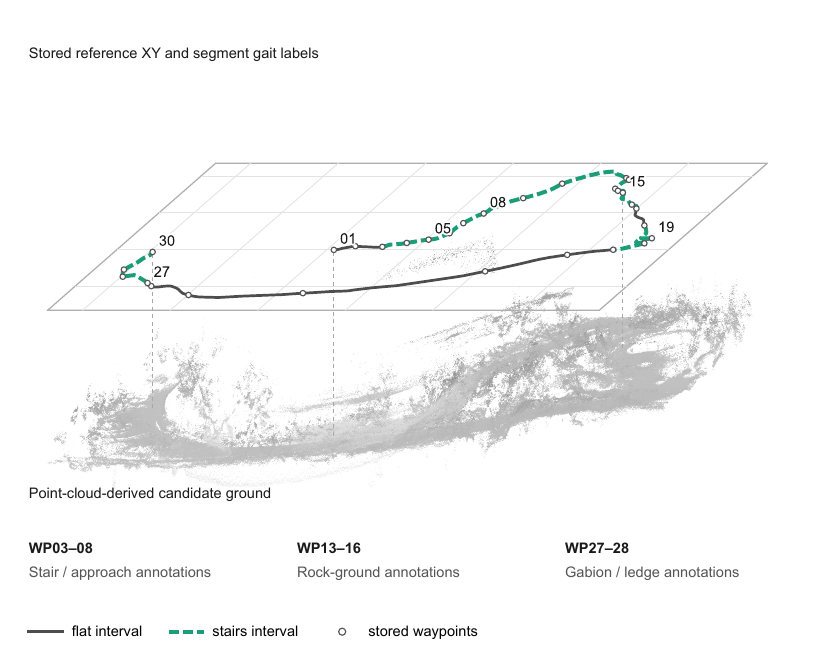}
\caption{\rev{\textbf{Map geometry and stored task information.} The view renders all
392,452 known 0.1 m cells of the candidate ground map in the display region. The upper plane
shows the archived simulation-reference XY, the waypoints, and the flat and stairs segment
labels. Terrain names come from stored annotations. The axonometric sketch combines these XY
task fields with map elevations and exaggerates height for readability.}}
\label{fig:taskmap}
\end{figure}

\section{Goal and route-build interfaces}\label{app:goalinterfaces}
For the field task, a waypoint is reached when some part of the robot intersects
a disk of radius $0.25\,\mathrm m$. Let $\mathcal B(q)$ be projected body geometry and
$W_i=\{p:\|p-w_i\|\le0.25\,\mathrm m\}$. Ordered completion requires
\begin{equation}
\exists\ t_1<\cdots<t_N:\quad \mathcal B(q(t_i))\cap W_i\ne\varnothing\quad\forall i.
\label{eq:goal}
\end{equation}
\rev{The simulation study also scores the stricter rule that the robot's centre comes within
$0.20\,\mathrm m$ of each goal}:
\begin{equation}
C_{0.20}=\mathbf 1[\exists\ t_1<\cdots<t_N:\ \|p_{\rm centre}(t_i)-w_i\|_2\le0.20\,\mathrm m\ \forall i].
\label{eq:strict}
\end{equation}
\rev{Separately, the field compiler places the line within $0.10\,\mathrm m$ of each mark
and adds a body-sweep proximity check. This compile-time geometric tolerance, the runtime gate
radius, and the sampled simulation score serve different purposes. Each evaluation fixes its
goal coordinates and their order before any method runs.}

\rev{When a waypoint is re-marked, we rebuild the bundle in the same registered map frame. The
compiler takes the latest passing mark, regenerates the route and manoeuvre fields, and reports
any waypoint without a usable mark. This is an explicit build step, and the new bundle must
pass the clearance checks and be reloaded before execution. The field route has 29 waypoints
and 28 segments. The simulation route has 30 waypoints.}

\rev{Before a gait switch, the field configuration requires the monitored translational and
yaw velocity channels to stay below a stillness threshold of $0.12$ for $0.3\,\mathrm s$. A
pose older than $0.3\,\mathrm s$ yields a zero command, and a grid or scan older than
$0.5\,\mathrm s$ is withheld from the planner, which stops the follower. The simulation
adapter records its acknowledgement model separately. All comparisons use the same handoff
procedure. The gait-interval state also ignores jitter at interval boundaries: a small backward
change in projected route progress keeps the current interval, and the field implementation
allows a retreat of up to $2\,\mathrm m$.}

\paragraph{Recorded crossing correspondence.}
\rev{Figure~\ref{fig:teaching} shows the whole compiled route and one stored local condition.
Near the crossing shown, a recorded pose lies 0.022 m from the compiled crossing, with a
heading of 2.107 rad against the stored approach direction of 2.089 rad. The bundle assigns a
stairs, a platform, and a stairs interval, in that order, to that region. The crossing position
and direction come from the local trails, with placement assisted by the map; the approach and
exit extents are configured. The stored rise field records the change in body height across
the passage. The example shows how recorded passage geometry and gait events become fields
that the executor uses; the following diagnostic tests their effect on approach behaviour.}

\begin{figure}[!htbp]
\centering\includegraphics[width=\linewidth]{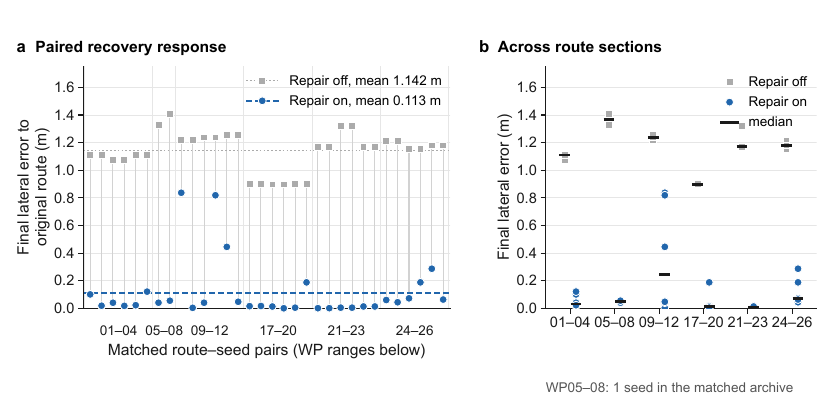}
\caption{\rev{\textbf{Rejoining the stored reference on six route segments.} Left: final
distance to the original reference in 32 matched rejoining-off/on pairs. Lines connect runs
with the same route, seed, and entry setting. Right: the same runs by segment, with median
bars. WP05--08 has one qualifying seed, and the other segments have three. The mean distance
falls from 1.142 m to 0.113 m. A pair enters the analysis only if all four factor combinations
received the displacement.}}
\label{fig:recoveryextension}
\end{figure}

\section{Recovery across route segments}\label{app:recoveryextension}
The expanded contact-simulation experiment crosses eight fixed segments (WP01--04,
05--08, 09--12, 13--16, 17--20, 21--23, 24--26, and 27--30), three seeds (101, 211,
and 307), rejoining off/on, entry conditioning off/on, and position displacement
absent/present. \rev{All runs use the live-sensing configuration and the pose-noise model of
the main protocol. Each simulation runs for at most 180 s. Of the 96 intended displacements,
69 are applied, 11 fail the ground-validity test, and 16 are not reached. The other 96
configurations are undisplaced controls. The outcome table below keeps all these groups.}

\begin{table}[ht]
\centering\small
\begin{tabular}{@{}llrr@{}}
\toprule
Rejoining & Entry conditioning & Control & Intended displacement\\
\midrule
Off & Off & 20.8 & 0.0\\
Off & On & 29.2 & 0.0\\
On & Off & 25.0 & 37.5\\
On & On & 45.8 & 50.0\\
\bottomrule
\end{tabular}
\caption{\rev{Ordered completion (\%) for the full expanded factorial. Each entry has 24
attempts: eight route segments times three seeds. The intended-displacement column divides by all 24
attempts with an intended displacement, including those where the ground check failed or the
trigger point was not reached.}}
\end{table}

\rev{For the paired analysis, a route--seed set qualifies only if all four combinations of
rejoining and entry setting receive the prescribed displacement. This gives 16 sets and 64
displaced executions, or 32 rejoining-off/on pairs. The pairs cover six segments: WP05--08
contributes one qualifying seed, and the other five contribute three each. Within each
route--seed set, the two entry settings are paired measurements, and we report matched counts
and distances. With rejoining enabled, the robot returns inside the original segment corridor
in all 32 matched cases; with it disabled, in none. Ordered completion is 59.4\% with rejoining
and 0\% without. A return counts only if the raw XY is first seen outside the frozen segment's corridor and later
inside it. Terminal distance projects the raw XY onto the original frozen centerline.
Figure~\ref{fig:recoveryextension} shows every matched terminal error, including runs that
fail later.}

{
\paragraph{Seed extension.} Entry conditioning in this appendix is the factor that
Section~\ref{sec:rejoin} calls the gait-entry lead. Twenty further seeds, the consecutive
primes from 401 to 521 and fixed before launch, repeat the same factorial with the same pinned
sources: 1,280 executions. We re-ran the 192 original executions on the new machine. They gave
the same completion outcome in 192 of 192 cases and reproduced every statistic above. Of the
640 intended displacements of the new seeds, 435 are applied, 73 fail the ground-validity test,
and 132 are not reached. In all, 97 route--seed sets qualify (20, 1, 16, 20, 20, and 20 for the
six contributing segments). On the new seeds alone, rejoining raises completion from 0\% to
56.7\% (interval of the difference 50.0 to 63.6 points). The supplementary material tabulates
all 1,280 executions, with every intended displacement in the denominator.

Over all 113 qualifying sets, the undisplaced runs with rejoining enabled complete 54.0\% (122 of
226), and the displaced runs of the same sets 57.1\% (129 of 226; difference $+3.1$ points, seed-block interval $-7.2$
to $+13.7$). All 226 displaced runs first return inside the corridor. Of the 97 that then do
not complete, 54 fall later on the segment, 42 reach its end without passing every goal centre
within 0.20 m in order, and one stalls. The 104 undisplaced runs that do not complete divide in
the same way (48 falls, 53 missed goals, 3 holds). The goals are missed by small margins. Of the
42 runs, 40 pass one goal outside the 0.20 m radius and two pass two goals outside it, by a
median of 1.8 cm and at most 6.9 cm, and 39 of them stay within 0.25 m of every goal, the
radius of the waypoint disks in the competition. The 53 undisplaced runs miss one goal each, by
a median of 1.3 cm and at most 6.4 cm, and 52 of them stay within 0.25 m. Scored at 0.25 m,
completion would be 74.3\% (168 of 226) after a displacement and 77.0\% (174 of 226) without
one. Rejoining also acts when no displacement is injected, because under pose noise the robot
stalls or leaves the line on its own. Over all 368 undisplaced pairs on the eight segments,
completion is 24.2\% with rejoining off and 44.8\% with it on (difference 20.7 points, interval
16.6 to 24.7), and 180 of the rejoining-off runs end in a hold, against 33 with rejoining on.

\begin{table}[!htbp]
\centering\small
\begin{tabular}{@{}lrr@{}}
\toprule
 & Previous (3 seeds) & All 23 seeds\\
\midrule
Qualifying route--seed sets / matched pairs & 16 / 32 & 113 / 226\\
Ordered completion, rejoining off & 0\% & 0\%\\
Ordered completion, rejoining on & 59.4\% & 57.1\%\\
\quad difference, 95\% seed-block bootstrap & & 50.9 to 63.4 points\\
Ordered completion, same sets undisplaced, rejoining on & 43.8\% & 54.0\%\\
Mean terminal distance, rejoining off / on & 1.14 m / 0.11 m & 1.10 m / 0.14 m\\
Returned inside corridor, rejoining off / on & 0 / 32 & 2 / 226\\
Entry conditioning after displacement, off / on & 56.3\% / 62.5\% & 59.3\% / 54.9\%\\
\quad difference, 95\% seed-block bootstrap & $-40.0$ to $+33.3$ & $-14.5$ to $+6.2$ points\\
Entry conditioning, undisplaced runs (184 sets) & & $+8.2$ points (3.0 to 13.3)\\
Recorded falls, rejoining off / on & 0 / 6 & 0 / 54\\
\bottomrule
\end{tabular}
\caption{\textbf{Recovery factorial with the seed extension.} Intervals resample complete
seeds 20,000 times. With rejoining enabled, the robot keeps going and can fall later. With
rejoining disabled, it holds.}
\label{tab:recovery23}
\end{table}

\paragraph{What the rejoining strategy contributes.} The rejoining factor compares a hold with a
track mode in which the robot walks forward while steering to a point on the persistent
reference. A third arm, fixed before launch, keeps the track mode but uses the tracker's
default recovery, the pursuit behaviour of our navigation core with its default settings: turn
toward the reference and side-step, with no forward walk. This arm ran on the same eight segments, 23
seeds, and both entry settings (368 displaced executions). The only source change is three
lines of the runner that select the arm, and a re-run of 64 existing cells with the patched
runner reproduced every outcome. Table~\ref{tab:rejoinarms} reports the 241 cells in which all
three arms received the displacement. Rejoining the reference in either form gives most of the
recovery (0\% to 44.4\%, interval of the difference 38.3 to 50.6 points). Walking forward while
steering adds 12.4 points (interval 2.9 to 22.4) and brings every run back inside the corridor.
Undisplaced runs of the same cells complete 55.2\% (133 of 241). The gap between the two
rejoining behaviours comes mainly from segments WP05--08 and WP17--20, where the side-step does
not bring the robot back to the line, and from WP21--23. On the robot, the side-step likewise
had no effect in the stairs gait, which is why we changed the behaviour.

\begin{table}[!htbp]
\centering\small
\begin{tabular}{@{}lrrrr@{}}
\toprule
Segment & Cells & Hold & Turn and side-step & Walk and steer\\
\midrule
WP01--04 & 46 & 0 & 33 & 32\\
WP05--08 & 19 & 0 & 0 & 10\\
WP09--12 & 38 & 0 & 18 & 18\\
WP17--20 & 46 & 0 & 3 & 18\\
WP21--23 & 46 & 0 & 28 & 38\\
WP24--26 & 46 & 0 & 25 & 21\\
\midrule
All: ordered completion & 241 & 0 (0\%) & 107 (44.4\%) & 137 (56.8\%)\\
Mean terminal distance & & 1.11 m & 0.28 m & 0.14 m\\
Returned inside corridor & & 2 & 222 & 241\\
Recorded falls & & 0 & 65 & 57\\
\bottomrule
\end{tabular}
\caption{\textbf{Three ways to respond to a displacement.} Completed runs per segment, over
the cells in which all three arms received the 1.2 m displacement. Intervals in the text
resample complete seeds 20,000 times.}
\label{tab:rejoinarms}
\end{table}

\paragraph{Displacement size.} The factorial displaces the robot by 1.2 m. A sweep fixed
before launch adds displacements of 0.3, 0.6, 0.9, 1.5 and 1.8 m on the six segments of the
paired analysis, with the same 23 seeds, both rejoining settings, and both entry settings. It
has 2,760 executions on the same machine as the runs above. Before the sweep, its driver
reproduced 144 of the existing 0 m and 1.2 m cells with identical trajectories. Local detours
stay inside the corridor, whose half-width is at most 0.8 m on this route. So a detour can
itself leave the robot 0.3 or 0.6 m off the line, and 0.9 m lies just beyond the
widest corridor. Table~\ref{tab:shiftsweep} and Figure~\ref{fig:shiftsweep} place the new
sizes next to the 0 m and 1.2 m cells.

Up to 1.2 m, 709 of the 711 rejoining runs that leave the corridor return inside it. At each
of these sizes, completion after rejoining lies within 3.4 points of the undisplaced runs of
the same sets. A robot that holds completes as often as an undisplaced one while the
displacement is smaller than the widest corridor half-width (34.6\% and 32.5\% at 0.3 and
0.6 m, against 34.2\% undisplaced on the same sets). It completes 1.7\% at 0.9 m and none
beyond. At 1.5 m the safety fallback of Section~\ref{sec:method} takes effect, which holds a
robot that is more than 1.5 m from the line. There, 104 of the 218 rejoining runs hold, and at
1.8 m all 156 do.

We repeated the whole sweep, with all 23 seeds, on a second machine (macOS, arm64): 3,864
executions. Individual outcomes differ between the two architectures because the contact
dynamics are chaotic, so this is an independent replication. On the second machine, completion
after rejoining is 53.7, 52.0, 52.9 and 55.6\% at 0.3 to 1.2 m, against 58.9 to 59.4\% for the
undisplaced runs of the same sets. The differences are $-5.3$, $-6.9$, $-6.2$ and $-3.8$ points,
and the seed-block interval contains zero at every size (at 1.2 m, $-13.9$ to $+5.9$). Of the
747 rejoining runs that leave the corridor, 745 return inside it. A robot that holds completes
43.9 and 30.5\% at 0.3 and 0.6 m, 4.1\% at 0.9 m, and none beyond. The two machines agree on the
returns and on the comparison with holding, and on neither does a displacement change
completion beyond the spread between seeds.

\begin{table}[!htbp]
\centering\small
\begin{tabular}{@{}lrrrrlr@{}}
\toprule
Displacement & Sets & Hold & Rejoin & Undisplaced & Rejoin $-$ undisplaced (95\%) & Returned / left\\
\midrule
0.3 m & 117 & 34.6 & 49.6 & 53.0 & $-3.4$ ($-11.0$ to $+3.9$) & 143 / 145\\
0.6 m & 117 & 32.5 & 53.4 & 53.0 & $+0.4$ ($-8.1$ to $+9.1$) & 136 / 136\\
0.9 m & 115 & 1.7 & 52.6 & 53.0 & $-0.4$ ($-9.5$ to $+8.4$) & 204 / 204\\
1.2 m & 113 & 0.0 & 57.1 & 54.0 & $+3.1$ ($-7.2$ to $+13.7$) & 226 / 226\\
1.5 m & 109 & 0.0 & 29.8 & 55.0 & $-25.2$ ($-33.9$ to $-16.7$) & 114 / 218\\
1.8 m & 78 & 0.0 & 0.0 & 56.4 & $-56.4$ ($-66.0$ to $-46.9$) & 0 / 156\\
\bottomrule
\end{tabular}

\vspace{6pt}
\begin{tabular}{@{}lrrr@{}}
\toprule
Reaching the next goal & Hold & Rejoin & Undisplaced\\
\midrule
0.3 m & 186 / 234 & 232 / 234 & 232 / 234\\
0.6 m & 187 / 234 & 234 / 234 & 232 / 234\\
0.9 m & 52 / 230 & 228 / 230 & 228 / 230\\
1.2 m & 2 / 226 & 226 / 226 & 224 / 226\\
1.5 m & 0 / 218 & 114 / 218 & 216 / 218\\
1.8 m & 0 / 156 & 0 / 156 & 154 / 156\\
\bottomrule
\end{tabular}
\caption{\textbf{Displacement-size sweep, 23 seeds.} Ordered completion (\%) on the route--seed
sets in which all four cells received the displacement, with a hold, with rejoining, and for
the undisplaced rejoining-on runs of the same sets (two runs per set and arm). Intervals
resample complete seeds 20,000 times. The last column gives, among the rejoining runs that
left the corridor, how many returned inside it. The 1.2 m row is the comparison of
Table~\ref{tab:recovery23}. The lower panel counts, on the same sets, the runs that reach the next
goal of the segment after the displacement (the window analysis at the end of this appendix).
The corridor half-width differs between segments (0.25 to 0.8 m), so some runs leave the corridor
at the smallest displacements.}
\label{tab:shiftsweep}
\end{table}

\begin{figure}[!htbp]
\centering\includegraphics[width=\linewidth]{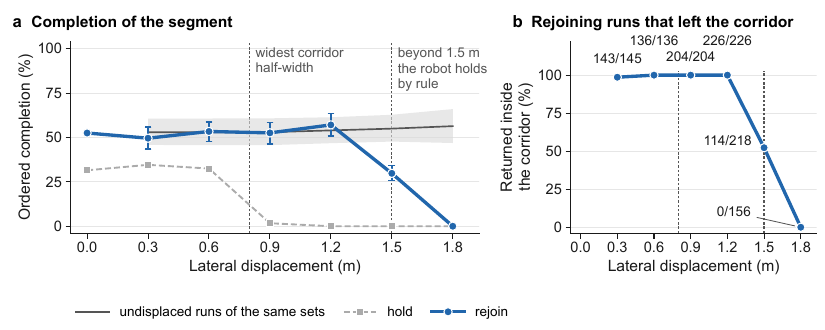}
\caption{\textbf{Recovery as a function of displacement size (23 seeds).} (a) Ordered
completion with rejoining and with a hold. The grey line and band show the undisplaced runs
of the same sets and their 95\% seed-block interval; the bars show the interval of the
rejoining rate. (b) Rejoining runs that returned inside the corridor, among those that left
it. The first dotted line marks the widest corridor half-width on the route; beyond the second, at 1.5 m, the robot is more than 1.5 m from the line and holds by rule.}
\label{fig:shiftsweep}
\end{figure}

\paragraph{Inserted obstacle.} A displacement stands in for what an object on the route causes.
We also placed the benchmark's test box on the route (0.45 m along the route, 0.50 m across,
0.55 m high, unknown to route preparation). We varied the lateral offset of its centre from the
route (0, 0.3, 0.6, and 0.9 m), with the 23 seeds, the four arms of the factorial, and no
displacement. Two sweeps, each fixed before launch, differ in where the box stands. The first
keeps the benchmark's position, at 5.0 m of route progress, on the benchmark's three segments
(1,104 executions). In the second, the box stands at
the midpoint of the first sub-segment that permits detours and is at least 6 m long. Four of
the six segments have such a sub-segment (1,472 executions; Table~\ref{tab:obstacle}).

On an open stretch, the robot goes around a box that blocks part of the corridor. With
rejoining, completion is 65.2\% and 60.9\% at 0.3 and 0.6 m, against 57.6\% with no box, and the
robot touches the box in 6 and 0 of 184 runs. Without rejoining, completion is 35.9\% and
34.8\%, against 33.7\% with no box, and 79 of 184 runs end in a hold at each offset, against 65
with no box. On WP24--26 the detour takes the robot to the edge of the corridor, and no run
without rejoining completes at these offsets (9 of 46 with no box). A box centred on the route
leaves no passage inside the 0.8 m corridor, and few runs complete (6.0\% and 9.2\%). There, the
robot without rejoining holds (161 of 184 runs). With rejoining, the stall triggers forward
rejoining; the robot touches the box in 154 runs and makes no progress.

The first sweep shows what happens where no detour is possible. At 5.0 m, the box stands
0.45 m past a waypoint on WP01--04 and on a staircase on WP05--08, where the stored route
permits no detour. Only on WP21--23 does it stand on an open stretch. On WP01--04, a robot
without rejoining holds in front of the box in all 46 runs at 0, 0.3, and 0.6 m. With
rejoining, it pushes on. Over the three segments it touches the box in 112, 94, and 61 of 138
runs at these offsets, and on the staircase all 46 runs at 0.3 m fall. On WP21--23 the robot
passes the box as in the second sweep (no contact in 92 runs without rejoining at 0.3 and
0.6 m, 2 with rejoining). In the benchmark, a stall in front of the box triggers the same
rejoining as any other stall.

\begin{table}[!htbp]
\centering\small
\begin{tabular}{@{}lrrlrrrr@{}}
\toprule
 & \multicolumn{2}{c}{Completion (\%)} & & \multicolumn{2}{c}{Touching the box} & \multicolumn{2}{c}{Ending in a hold}\\
Box offset & off & on & On $-$ no box (95\%) & off & on & off & on\\
\midrule
no box & 33.7 & 57.6 & & 0 & 0 & 65 & 2\\
0 m & 6.0 & 9.2 & $-48.4$ ($-58.2$ to $-38.6$) & 6 & 154 & 161 & 36\\
0.3 m & 35.9 & 65.2 & $+7.6$ ($-4.9$ to $+19.0$) & 0 & 6 & 79 & 0\\
0.6 m & 34.8 & 60.9 & $+3.3$ ($-4.9$ to $+12.0$) & 0 & 0 & 79 & 2\\
0.9 m & 39.7 & 57.6 & $0.0$ ($-11.4$ to $+11.4$) & 0 & 0 & 65 & 0\\
\bottomrule
\end{tabular}
\caption{\textbf{Inserted box on open stretches, 23 seeds.} Ordered completion, runs that
touch the box, and runs that end in a hold, with rejoining off and on: 184 runs per arm and
offset (four segments, two entry settings). The offset is the lateral distance of the box
centre from the route. The corridor half-width is 0.8 m, and the body is 0.6 m wide. Intervals
resample complete seeds 20,000 times. The no-box row gives the undisplaced runs of the same
segments from the factorial.}
\label{tab:obstacle}
\end{table}

\paragraph{Scoring by the competition's rule.} The tables above use the benchmark's
rule: the robot's centre must come within 0.20 m of every goal in order. In the
competition, a waypoint counted when the body touched its 0.25 m disk. We re-scored the
recorded runs by that rule from the logged poses, with the $0.9\times0.5$ m body footprint of
the navigation stack. As before, goals must be touched in order, and a fall voids the run. No
simulation was run for this. To check the reconstruction, we applied the same procedure to the centre at
0.20 m. It agrees with the simulator's own completion flag in 1,835 of 1,840 runs.
Table~\ref{tab:comprule} gives both readings. After a 1.2 m displacement, rejoining runs
complete 75.7\% under the competition's rule (interval of the difference to holding 70.5 to
80.2 points), against 77.4\% for the same sets undisplaced (difference $-1.8$ points, interval
$-10.0$ to $+7.0$). The 55 runs that do not complete are 54 falls and one stall, and the 42
runs that pass a goal just outside 0.20 m all count. Over 0.3 to 1.2 m, the rejoining runs
complete 75.2, 77.4, 76.5, and 75.7\%, within 1.8 points of the undisplaced runs of the same
sets (76.9 to 77.4\%). A robot that holds completes 54.3, 53.0, 3.9, and 0.4\%. The gait-entry
lead changes completion by $+6.2$ points on undisplaced runs (interval $-0.5$ to $+13.0$) and by
$+0.9$ points on displaced runs ($-9.6$ to $+11.6$). With a box on an open stretch, rejoining
runs complete 85.3, 81.5, and 85.9\% at 0.3, 0.6, and 0.9 m, against 84.8\% with no box (50.0,
48.4, 57.1, and 53.8\% without rejoining).

\begin{table}[!htbp]
\centering\small
\begin{tabular}{@{}lrr@{}}
\toprule
Ordered completion & Centre within 0.20 m & Body touches the 0.25 m disk\\
\midrule
1.2 m displacement, hold (226 runs) & 0.0\% & 0.4\%\\
1.2 m displacement, rejoining & 57.1\% & 75.7\%\\
Same sets, not displaced & 54.0\% & 77.4\%\\
\midrule
Three behaviours, 241 cells: hold & 0.0\% & 0.4\%\\
\quad default turn and side-step & 44.4\% & 61.4\%\\
\quad walking forward while steering & 56.8\% & 75.9\%\\
\quad difference of the last two (95\%) & 12.4 (2.9 to 22.4) & 14.5 (8.2 to 21.3)\\
\quad same cells, not displaced & 55.2\% & 78.0\%\\
\midrule
Nothing injected, 368 pairs: rejoining off & 24.2\% & 39.9\%\\
\quad rejoining on & 44.8\% & 63.6\%\\
\quad difference (95\%) & 20.7 (16.6 to 24.7) & 23.6 (19.0 to 28.3)\\
\bottomrule
\end{tabular}
\caption{\textbf{The recovery results under the two scoring rules.} Left: the benchmark's rule, used in
Tables~\ref{tab:recovery23} to~\ref{tab:obstacle}. Right: the
competition's waypoint rule, used in Section~\ref{sec:rejoin} and Figure~\ref{fig:overview}.
Intervals resample complete seeds 20,000 times.}
\label{tab:comprule}
\end{table}

\paragraph{From the event to the next goal.} Ordered completion scores a whole segment of three
or four goals, so it also counts what happens several terrain transitions after the event. To
measure the response to the event itself, we score the same runs on a window from the event to
the next goal of the segment, using the benchmark's own record of ordered hits at 0.20 m
(definitions in the supplementary material). For a displacement, the window opens when the
displacement is applied. For a box, it opens 3.5 m before the box, the planner's horizon.
Undisturbed runs are scored from the same point of the route.

After a 1.2 m displacement, all 226 rejoining runs reach the next goal, against 2 of the 226
that hold and 224 of the 226 undisplaced runs of the same sets (Table~\ref{tab:shiftsweep},
lower panel). Over 0.3 to 1.2 m, the counts are 920 of 924 with rejoining, 427 of 924 with a
hold, and 916 of 924 undisplaced. On the second machine they are 958, 463, and 960 of 968 (all 23 seeds). Each of the 97 displaced runs that do not complete first reached the next goal,
and so did 102 of the 104 undisplaced ones. Completion is lost later on the segment, with or
without a displacement. At 1.5 m, 114 of 218 rejoining runs reach the next goal. At 1.8 m none
does, because of the 1.5 m rule. On the three original seeds alone, the counts at 1.2 m are 32
of 32 with rejoining and 0 of 32 with a hold. For the three recovery behaviours of
Table~\ref{tab:rejoinarms}, the next goal is reached in 2 of the 241 cells with the hold, in 216
with the default behaviour, and in all 241 with walking forward while steering (difference
between the last two: 10.4 points, seed-block interval 7.8 to 13.1). The 25 misses of the
default behaviour are all 19 cells of WP05--08, four of WP24--26, and two of WP09--12. Take the
cells where both the displaced and the undisplaced run reach the next goal. There, the time from
the event to the next goal is a median of 1.6 s longer after the displacement with walking
forward while steering (quartiles 0.7 to 2.1 s, 224 cells), and 5.8 s longer with the default
behaviour (4.3 to 7.8 s, 216 cells). Over the whole segment, where both runs complete, the
medians are 2.0 s and 6.3 s (72 and 57 cells).

With a box on an open stretch, the rejoining runs reach the next goal without touching the box
in 96.7\%, 98.9\%, and 100\% of 184 runs at 0.3, 0.6, and 0.9 m, against 98.9\% with no box. At
0.3 m all 184 reach it, and six touch the box. Without rejoining, the rates are 71.7\%, 75.0\%,
and 83.7\%, against 82.6\%. At these offsets, the two arms differ only on WP24--26. There, no run
without rejoining reaches the next goal at 0.3 and 0.6 m (14 of 46 with no box), while 46 and
44 of 46 do with rejoining. On the other three segments, both arms reach it in at least 42 of 46
runs. A box centred on the route is passed in 13.0\% and 16.3\% of runs. In the first sweep,
the box stands at a waypoint or on a staircase on two of the three segments. In that sweep, the
rates with rejoining are 18.8\%, 31.9\%, 55.8\%, and 99.3\% at 0, 0.3, 0.6, and 0.9 m, against
100\% with no box. Without rejoining they are 17.5\%, 32.1\%, 33.6\%, and 85.4\%, against
85.4\%. One run ends before the window opens and is not counted.
}

\begin{figure}[!htbp]
\centering\includegraphics[width=\linewidth]{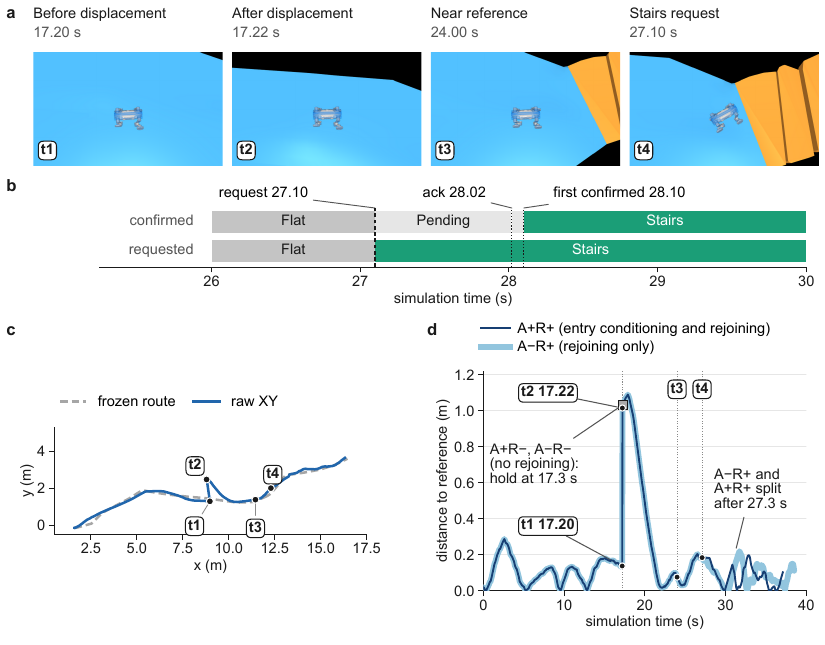}
\caption{\rev{\textbf{A displaced robot rejoins the reference and continues toward the next
transition.} The four replay frames, the XY trajectory, and the event markers come from one
earlier nominal run with rejoining on and entry conditioning on. The distance curves compare
all four matched settings (A: entry conditioning; R: rejoining) and project the raw XY onto the
frozen original centerline. Curves that overlap are drawn without offsets. The strips of
requested and confirmed gait zoom into 26--30 s of the run shown; the last replay frame comes
before the acknowledgement. The replays render stored simulator states at fixed simulation
times, with a robot-centred camera at a shared view direction and distance.}}
\label{fig:rejoin}
\end{figure}

\paragraph{Earlier nominal mechanism diagnostic.}
\rev{At 8 m of route progress, we displace the robot 1.2 m sideways, after the existing
ground-validity check. Of the 36 planned cells (three routes, three seeds, four factor
combinations), 24 receive the displacement, on two routes. The third route fails the validity check in all 12 cells and is used as an undisplaced control. In the displaced cells, the robot
completes 100\% with rejoining and 0\% without it. Within each rejoining setting, both entry
settings have the same completion outcome. All 12 undisplaced controls complete. The replay in
Figure~\ref{fig:rejoin} comes from this nominal diagnostic. On one of the two displaced routes,
all settings stay in the flat gait throughout, so that segment shows recovery to the reference
with no gait change.}

\section{Fixed locomotion actor interface}\label{app:actorinterface}
\paragraph{Temporal inputs to the fixed locomotion actors.}
In the simulation adapter, the active flat or stair actor has the interface
\begin{equation}
 a_t=\pi_{\theta_{g_t}}
 \bigl(z_t,a_{t-1},\sin\phi_t,\cos\phi_t\bigr),
 \label{eq:fixed_actor_temporal_inputs}
\end{equation}
where $g_t$ denotes the active actor, $z_t\in\mathbb R^{41}$ contains scaled
body angular velocity, projected gravity, the velocity command, relative joint
positions, and joint velocities, and $a_t\in\mathbb R^{16}$ is its raw action.
\rev{The previous action and the gait phase are explicit temporal inputs. The wrapper resets
both when an actor takes over. It advances the flat actor's phase on a fixed clock, and the
stair actor's phase when the velocity command exceeds its activation threshold. The parameters
$\theta_g$ stay fixed during the navigation comparisons. This actor interface is separate from
the persistent route reference $L$ and from the runtime navigation state $S_t$. The physical
robot uses a separate factory gait stack.}

\paragraph{Offline temporal-input diagnostic.}
\rev{We checked whether the two frozen simulation actors respond to their explicit temporal
inputs. For each actor, we took 16 equally spaced recorded states from the interval in which
that actor was active in one completed rollout. The saved trace does not contain the previous
actions and phases, so we built inputs with zero previous action and four prescribed phases.
Central differences perturbed each previous-action coordinate by $\pm0.05$ and the phase by
$\pm0.05$ radians. In both actors, the action responded to each input group with nonzero
sensitivity. In this diagnostic, querying the same base inputs again after other calls gave
identical outputs. The measurements describe the action sensitivity conditional on these
prescribed temporal inputs.}

\rev{The 128 controlled observations combine 32 recorded body states with four phases and are
queried offline. For the flat and stair actors, the median spectral norms of the scaled
leg-target Jacobian with respect to the previous action are 0.2172 and 0.2154 rad. The
corresponding finite-difference norms for the phase are 0.1096 and 0.1538 rad/rad. The logs do
not contain the actual previous actions and phases, so these constructed inputs can lie outside
the closed-loop observation distribution.}

\clearpage
\section{\rev{Entry conditions of our learned locomotion policies}}\label{app:learnedskills}
\begin{figure}[!htbp]
\centering\includegraphics[width=0.92\linewidth]{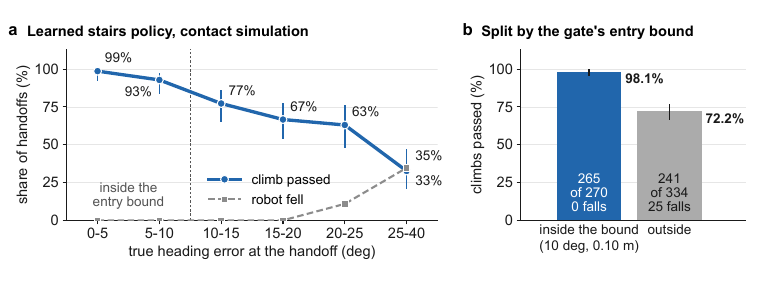}
\caption{\rev{\textbf{The learned stairs policy succeeds from a limited set of entry states.}
Inside the gate's bound 98.1\% of entries pass, and outside it 72.2\%. Contact simulation, 378
runs. Left: the share of handoffs that pass and the share that end in a fall, against the true
heading error at the handoff; bars are 95\% intervals. Right: all 604 handoffs, split by the
gate's entry bound.}}
\label{fig:entrysweep}
\end{figure}

\begin{figure}[!htbp]
\centering\includegraphics[width=\linewidth]{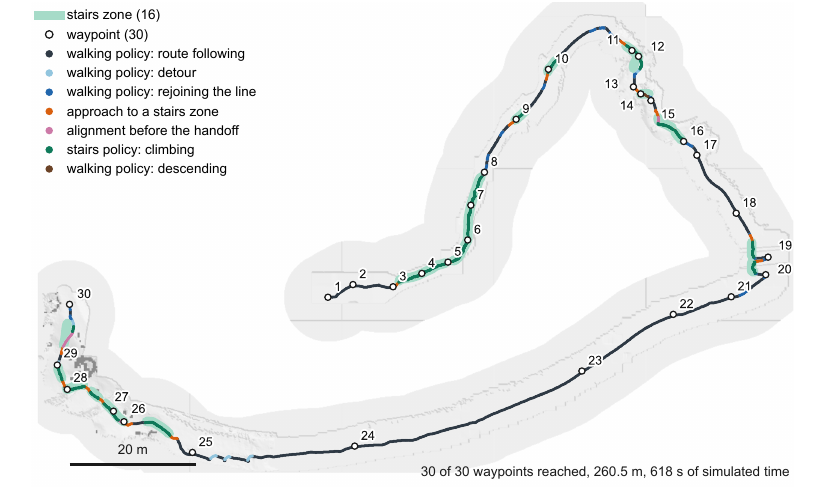}
\caption{\rev{\textbf{A full-route run in contact simulation with our two learned policies.} One
recorded run of the archived stack: 30 of 30 waypoints over 260.5 m in 618 s of simulated time.
The stairs policy drives the green intervals in 15 of the 16 stairs zones, and the walking
policy drives the rest, including the descent through the remaining zone (brown). Every stairs
zone is preceded by an approach (orange) or alignment (pink) phase.}}
\label{fig:fullroute}
\end{figure}

\rev{The two simulation actors are policies that we trained with PPO: a walking policy and a
stairs policy, each with 59 observations. They are frozen in every experiment of this paper.
Before the factory gait interface was available, the navigation stack drove these two
policies in contact simulation over the full 260 m route, with 30 waypoints and 16 stairs
zones. The archived batch from that period shows the premise of this paper with learned skills.

\paragraph{Measured entry limits.} We ran synthetic tests on the two policies, treating them as
black boxes. These tests gave the capability profile that the archived stack loads; that stack
predates the event interface of Section~\ref{sec:method}. The stairs policy climbs steps of up
to 0.18 m and cannot turn on stairs. At entry it tolerates at most $25^\circ$ of skew to the
first riser and 0.2 m of lateral error. The walking policy tolerates $30^\circ$ and 0.1 m, and
it stalls on small steps below 0.5 m/s. We could run such tests because these policies are ours
and run in simulation; we have no such tests for the factory gaits. A third policy, trained on a
0.38 m ledge, was carried on the robot as a backup to the factory ledge gait; the robot's router could hand
it control only inside an explicit entry window: 0.62 to 0.70 m from the edge, within 0.25 m
laterally and $6^\circ$ in heading, at 0.08 to 0.20 m/s, and turning at less than 0.10 rad/s.
The navigation had to deliver the robot inside this window.

\paragraph{Alignment before every stairs zone.} Each zone stores where the approach
starts and the direction of the route at the first step. The stack slows 1.2 m before the
zone, pivots to the stored direction with a tolerance of $6^\circ$, and only then hands
control to the stairs policy. It logs the heading and lateral error at every handoff.
\rev{Figure~\ref{fig:fullroute} shows one recorded full-route run.}

\paragraph{What the archive shows.} The archive has sixty end-to-end runs: 32 seeds without
disturbance and 28 with localization bias, localization noise, shoves, or added obstacles. They
contain 655 handoffs to the stairs policy, in 15 of the 16 zones. The heading error at the
handoff is within $6^\circ$ in 592 of them and never above $15.6^\circ$, and 632 climbs
(96.5\%) pass. The 23 that do not pass cluster where the stored route crosses the first riser
at an angle (Table~\ref{tab:learnedskills}). They are 17 of 181 climbs in the four zones with
$20^\circ$ to $50^\circ$ of skew, against 6 of 474 in the other eleven zones (Fisher exact
$p=3\times10^{-6}$). Eight of the ten falls are in those four zones. The arrival state matters
as well. Started from rest at the waypoint before the hardest of these staircases, the same
policies pass it in 59 of 59 isolated runs. When the robot arrives from the preceding segments,
48 of 57 climbs of that zone pass.

\paragraph{End to end.} An earlier controller without recovery completed the full route in
4 of 24 seeds. We then added recovery, which returns the robot to the stored route when it
measures a deviation. With recovery, 19 of 32 seeds completed, and 14 of 28 runs completed
under the seven disturbance types.

\paragraph{A controlled sweep of the entry error.} The archive links failures to oblique
entries, but its zones also differ in other ways. A controlled sweep isolates the entry error.
We ran 378 further contact simulations on the archived stack, with no source change, on three
stairs segments that the stored route enters nearly square (five zones, steps of 0.09 to
0.15 m), with six seeds each. We add a constant bias to the pose that the navigation receives.
The stack then aligns using a wrong pose estimate, so the stairs policy takes over with a true
heading or lateral error that the stack does not see, and climbs with it. The bias is up to
$30^\circ$ of heading in $5^\circ$ steps, or 0.1 to 0.4 m of lateral offset. We measure the
true error at each handoff against the unbiased run of the same segment and seed. The bins, the
outcome rule, and the exclusions were fixed before launch, and all 378 runs are kept. They form
one batch on one machine and are not pooled with the archive.

Figure~\ref{fig:entrysweep} and Table~\ref{tab:entrysweep} give the 604 handoffs. The pass
rate of the learned stairs policy falls monotonically with the heading error at entry, from
98.7\% below $5^\circ$ to 32.7\% beyond $25^\circ$. Below $20^\circ$ no entry falls; the
entries that do not pass miss a waypoint and trigger recovery. Falls begin at $20^\circ$
(5 of 46) and reach 17 of 49 beyond $25^\circ$, in line with the $25^\circ$ skew limit of
the capability profile. Lateral error up to 0.4 m changes the pass rate little (89.1\% in the
weakest bin, with three falls between 0.2 and 0.3 m). On these staircases, heading is the entry
variable that matters. Used as a classifier, the gate's entry bound ($10^\circ$, 0.10 m)
separates a pass rate of 98.1\% inside it (265 of 270; 95\% interval 95.7 to 99.2; no fall)
from 72.2\% outside (241 of 334; 67.1 to 76.7; 25 falls). Unperturbed entries pass at 96.7\%
(29 of 30). Two of the five zones carry the effect. One is the 1.7 m flight with 0.15 m steps
(35 of 37 inside the bound against 35 of 89 outside), and the other is a staircase with 0.11 m
steps (32 of 35 against 59 of 91). The other three zones pass at 90\% or more outside the
bound, and the lowest steps (0.09 m) pass all 126 entries.

Together, the archive and the sweep establish, in contact simulation and with learned
policies, the two facts that the method rests on. A frozen learned skill succeeds from a
limited set of entry states. Inside a bound of the size that the gate enforces, it passes
98.1\% of entries with no fall, against 72.2\% outside.}

\rev{\paragraph{The alignment switched on and off.} The sweep sets the entry error directly. A
further batch of 288 contact simulations tests the check itself in a matched design; it is the
contact-physics counterpart of the factorial in Section~\ref{sec:isolate}. It uses the same three
segments and five zones, with twelve seeds. When the robot is 0.76 m before a zone, the distance
used in the factorial, we displace it sideways by 0.2, 0.4 or 0.6 m. In one arm the stack runs as
archived and aligns to the stored direction before each zone. In the other, the stairs policy
takes over at the scheduled arc length whatever the pose; this is gait scheduling without an
entry check. No source file is changed, everything else is identical, and undisplaced runs of
both arms serve as controls. The protocol and the analysis were fixed before launch, and all runs
are kept. With the alignment, all 180 displaced climbs pass and none falls. Without it, 171 pass
and the other nine fall (Table~\ref{tab:e4}; difference 5.0 points, 95\% interval 1.7 to 8.9
from resampling seeds; Fisher exact $p=0.0035$). The alignment lowers the median heading error
at the handoff from $28.1^\circ$ to $5.2^\circ$ and costs 1.6 s per zone. The stored direction
matters most where the approach is sharpest. In the zone that the robot reaches at about
$50^\circ$ without the alignment, 9 of 36 climbs fall, against none with it ($p=0.002$). In the
other four zones, the policy also climbs from entries of up to $35^\circ$ that the stack sees.
In the sweep, by contrast, a pose bias hidden from the stack keeps the error through the climb.}

\rev{\paragraph{The ledge policy at the competition's ledge.} The factorial of
Section~\ref{sec:isolate} scores ledge entry geometry kinematically. To see what entry means in
contact physics at a ledge, we placed the ledge policy, unchanged and configured as on the
robot, before the gabion ledge of the reconstructed competition scene (WP27 to WP28). This ledge
differs from the one the policy was trained on. It is 0.32 m high, with a sloped approach and an
uneven lip. A full grid of entry states covers 0.55 to 1.2 m from the lip, lateral offsets of up
to 0.3 m, headings of up to $30^\circ$ to either side, and speeds of 0.1 to 0.3 m/s, with three
seeds: 2,925 entries, all kept (Table~\ref{tab:ledgesweep}). Falls rise with the heading error,
from 8.9\% within $5^\circ$ to 18.4\% at $30^\circ$. Used as a classifier, the gate's bound
(10 cm, $10^\circ$) separates 5.3\% falls inside it (12 of 225) from 12.3\% outside (333 of
2,700; Fisher exact $p=0.001$). The success rate differs less (61.8\% against 56.3\%,
$p=0.12$). About a third of the entries on either side stall at the lip within the 20 s limit.
A stall leaves the robot upright; a fall ends the run.}

\begin{table}[!htbp]
\centering\small
\begin{tabular}{@{}lrr@{\qquad}lrr@{}}
\toprule
Heading error at handoff & Passed & Rate & Lateral error at handoff & Passed & Rate\\
\midrule
$0$ to $5^\circ$ & 74 / 75 & 98.7\% & 0 to 0.05 m & 52 / 52 & 100\%\\
$5$ to $10^\circ$ & 65 / 70 & 92.9\% & 0.05 to 0.10 m & 71 / 71 & 100\%\\
$10$ to $15^\circ$ & 51 / 66 & 77.3\% & 0.10 to 0.20 m & 52 / 53 & 98.1\%\\
$15$ to $20^\circ$ & 42 / 63 & 66.7\% & 0.20 to 0.30 m & 41 / 46 & 89.1\%\\
$20$ to $25^\circ$ & 29 / 46 & 63.0\% & 0.30 to 0.60 m & 13 / 13 & 100\%\\
$25$ to $40^\circ$ & 16 / 49 & 32.7\% & & & \\
\midrule
Inside the gate's bound & 265 / 270 & 98.1\% & Outside the bound & 241 / 334 & 72.2\%\\
\bottomrule
\end{tabular}
\caption{\rev{\textbf{Pass rate of the learned stairs policy against the true entry error
(contact simulation, 378 runs, 604 handoffs).} Left: runs with a heading bias. Right: runs
with a lateral bias. Bottom: all handoffs, split by the gate's entry bound of $10^\circ$ and
0.10 m. A handoff passes when the stairs zone is completed and control returns to the walking
policy.}}
\label{tab:entrysweep}
\end{table}

\begin{table}[!htbp]
\centering\small
\begin{tabular}{@{}lrrrrr@{}}
\toprule
Stairs zones & Zones & Handoffs & Passed & Falls & Recover or hold\\
\midrule
First riser crossed at $20^\circ$ to $50^\circ$ & 4 & 181 & 164 (90.6\%) & 8 & 9\\
First riser crossed below $10^\circ$, or no riser & 11 & 474 & 468 (98.7\%) & 2 & 4\\
\midrule
All & 15 & 655 & 632 (96.5\%) & 10 & 13\\
\bottomrule
\end{tabular}
\caption{\rev{\textbf{Climbs by the learned stairs policy in 60 archived end-to-end
contact-simulation runs.} A handoff passes when the stairs zone is completed and control
returns to the walking policy. ``Recover'' means that a waypoint inside the zone was missed and
the recovery behaviour was triggered.}}
\label{tab:learnedskills}
\end{table}

\begin{table}[!htbp]
\centering\small
\begin{tabular}{@{}lrr@{}}
\toprule
Displaced 0.76 m before each zone & Alignment on & Alignment off\\
\midrule
Climbs passed & 180 / 180 (100\%) & 171 / 180 (95.0\%)\\
Falls & 0 & 9\\
Median heading error at the handoff & $5.2^\circ$ & $28.1^\circ$\\
Handoffs within $10^\circ$ & 180 / 180 & 37 / 180\\
Zone reached at about $50^\circ$ without alignment & 36 / 36 & 27 / 36\\
Other four zones & 144 / 144 & 144 / 144\\
Median time from displacement to handoff & 3.8 s & 2.2 s\\
Segment completed in order (0.20 m) & 77.8\% & 75.0\%\\
\midrule
Undisplaced controls: climbs passed & 57 / 60 & 57 / 60\\
\bottomrule
\end{tabular}
\caption{\rev{\textbf{The alignment before each stairs zone switched on and off (contact
simulation, learned stairs policy, 288 runs, twelve seeds).} The robot is displaced sideways by
0.2 to 0.6 m when it is 0.76 m before each of five zones. Without alignment the stairs policy
takes over at the scheduled arc length whatever the pose. In the zone reached at about $50^\circ$
the difference has Fisher exact $p=0.002$.}}
\label{tab:e4}
\end{table}

\begin{table}[!htbp]
\centering\small
\begin{tabular}{@{}lrrrr@{}}
\toprule
Heading error at entry & Entries & Climbed & Fell & Stalled\\
\midrule
$0$ to $5^\circ$ & 675 & 57.9\% & 8.9\% & 33.2\%\\
$10$ to $15^\circ$ & 900 & 60.8\% & 9.9\% & 29.3\%\\
$20$ to $25^\circ$ & 900 & 55.0\% & 12.6\% & 32.4\%\\
$30^\circ$ & 450 & 50.4\% & 18.4\% & 31.1\%\\
\midrule
Inside the gate's bound (10 cm, $10^\circ$) & 225 & 61.8\% & 5.3\% & 32.9\%\\
Outside the bound & 2{,}700 & 56.3\% & 12.3\% & 31.3\%\\
All entries & 2{,}925 & 56.8\% & 11.8\% & 31.5\%\\
\bottomrule
\end{tabular}
\caption{\rev{\textbf{The learned ledge policy at the competition's gabion ledge (contact simulation,
2,925 entries).} Entries start 0.55 to 1.2 m from the lip at 0.1 to 0.3 m/s, with lateral offsets of
up to 0.3 m. A climb counts when all four wheels are on the deck and the robot stays upright for 1 s. A
run that neither climbs nor falls within 20 s counts as a stall.}}
\label{tab:ledgesweep}
\end{table}

\section{Controller-training configuration}\label{app:archivedtraining}
\begin{figure}[!htbp]
\centering\includegraphics[width=\linewidth]{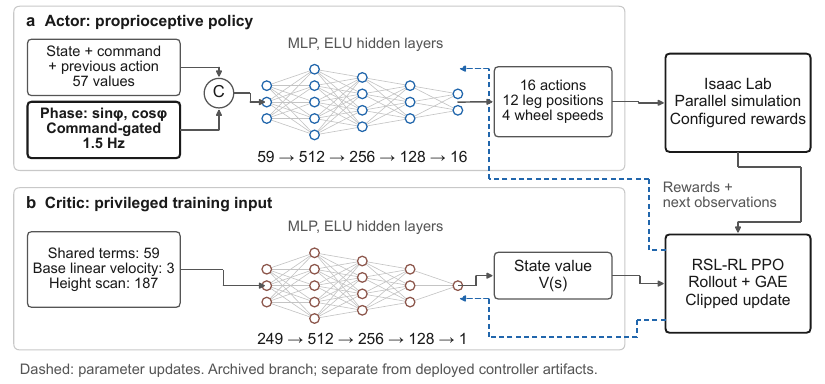}
\caption{\rev{\textbf{Asymmetric actor--critic in the archived controller-training branch.}
The critic receives privileged training information. The actor uses its configured
proprioceptive, command, previous-action, and phase inputs. The numbers give the hidden
widths; the drawn neurons are schematic. Solid arrows show data flow, and dashed arrows show
parameter updates. The actor and critic use matching observation terms with different scaling
and noise settings. The diagram describes the archived controller-training setup.}}
\label{fig:archivedppo}
\end{figure}

\rev{The project also contains a PPO training branch in Isaac Lab and RSL-RL with an
asymmetric actor--critic, shown in Figure~\ref{fig:archivedppo}. The actor receives 59 values,
including the previous actions and two phase terms. The critic receives 249 values; it adds the
linear velocity and a 187-value height scan. Both networks have hidden widths of 512, 256, and
128 with ELU activations. The 16 action channels set twelve leg-position targets and four
wheel-speed targets. This configuration belongs to a separate archived training branch.} \rev{The stairs actor
evaluated in this paper is checkpoint 1150 of a run with this contract (59 actor inputs,
249 critic inputs, 16 actions); Appendix~\ref{app:learnedskills} describes how the two
actors were used before the factory gaits. The factory gait stack on the robot is separate
and was not trained by us.}

\clearpage
\section{Expanded multi-terrain route comparison}\label{app:multiterrain}
\begin{figure}[!htbp]
\centering\includegraphics[width=\linewidth]{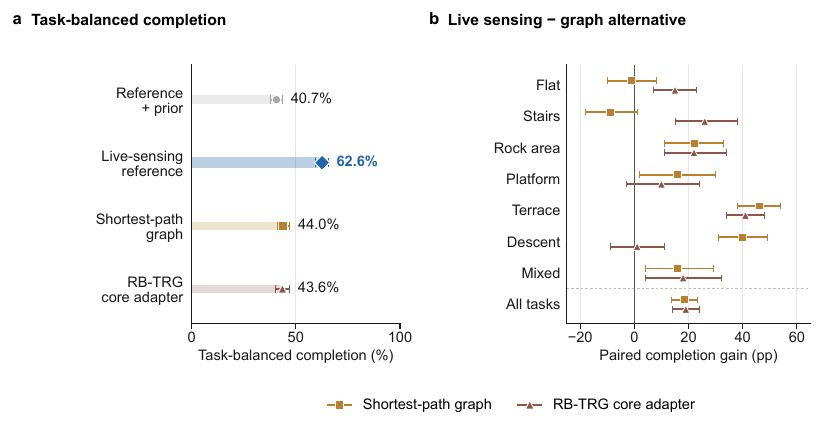}
\caption{\rev{\textbf{Live sensing raises aggregate completion on the fixed task suite.}
(a) Task-balanced completion over 700 trials per configuration. Each task has weight $1/7$, and
each of the two conditions weight $1/2$. (b) Paired differences against both graph
alternatives, by task and pooled. The 95\% intervals come from 20,000 percentile seed-block
bootstrap draws, which keep both conditions and all task strata together within each seed.
They describe seed variability on one reconstructed scene. Appendix~\ref{app:multiterrain}
gives the task protocol, the corrected rock-area start, and the full completion and fall
counts.}}
\label{fig:multiterrain-expanded}
\end{figure}
\rev{The final cohort combines 20 initial and 30 further seeds.} For configuration
$m$, the reported task-balanced completion is
\begin{equation}
\widehat R_m=\sum_{t=1}^{7}\frac{1}{7}
\sum_{c=1}^{2}\frac{1}{2}
\left(\frac{1}{50}\sum_{s=1}^{50}y_{m,t,c,s}\right),
\end{equation}
where $y$ indicates ordered completion. \rev{Tasks and conditions have equal weights. Every
stratum has the same number of trials, so this equals the pooled completion rate. Differences
are $100(\widehat R_{\mathrm{live}}-\widehat R_m)$ in percentage points. Each seed is paired
across seven fixed subroutes, the nominal and pose-noise conditions, and all four
configurations. The subroutes are fixed parts of one reconstructed scene and can overlap.
Execution and ordered-goal criteria are the same as in the main protocol. Confidence intervals
resample complete seed blocks with replacement 20,000 times and use the 2.5th and 97.5th
percentiles. Paired differences are computed within task, condition, and seed before the blocks
are resampled. The intervals measure seed variability for these tasks and settings.}

\paragraph{Rock-area initialization.}
\rev{The first start, at WP11, placed the robot in an inverted configuration. All 160 attempts
with that setup are kept in a separate ledger of invalid setups. The corrected task starts
earlier, at WP10, and includes the approach to the rock region. Its 160 initial runs and 240
further runs are part of the final 2,800 valid trials. All other failure outcomes are kept. We
finished the trajectory scoring and the file-integrity audit before assembling the final
plots.}

\begin{table}[ht]
\centering\small
\begin{tabular}{@{}lrrrrrrrr@{}}
\toprule
& \multicolumn{2}{c}{Reference + prior} & \multicolumn{2}{c}{Live sensing} & \multicolumn{2}{c}{Shortest path} & \multicolumn{2}{c}{RB-TRG core}\\
Task & C & F & C & F & C & F & C & F\\
\midrule
Flat & 87 & 0 & 87 & 0 & 88 & 0 & 72 & 0\\
Stairs & 18 & 72 & 67 & 16 & 76 & 10 & 41 & 56\\
Rock approach & 58 & 36 & 30 & 63 & 8 & 91 & 8 & 89\\
Platform & 45 & 42 & 58 & 23 & 42 & 46 & 48 & 44\\
Terrace & 5 & 91 & 58 & 34 & 12 & 84 & 17 & 80\\
Descent & 22 & 67 & 74 & 13 & 34 & 53 & 73 & 11\\
Mixed & 50 & 42 & 64 & 23 & 48 & 46 & 46 & 44\\
\midrule
Total & 285 & 350 & 438 & 172 & 308 & 330 & 305 & 324\\
\bottomrule
\end{tabular}
\caption{\rev{Full raw counts for the final route comparison. C: ordered completion; F:
recorded fall. Each task has 100 trials per configuration (50 seeds times two conditions), and
each configuration has 700 trials in total. Falls and goal completion are scored
separately.}}
\label{tab:final50counts}
\end{table}

\section{Vertical geometry stress test}\label{app:geometry}
\rev{A separate host runs ten paired pose-noise seeds on four tasks at each vertical scale,
including an unscaled control. Scene heights change, while horizontal geometry and robot
parameters stay fixed. The descent task includes an approach waypoint and differs from the
earlier multi-terrain descent subroute. All 640 runs are kept. Aggregate completion is 25.6\%
for the prior-enabled reference, 27.5\% for live sensing, 38.8\% for shortest-path planning, and
27.5\% for the RB-TRG adapter. The test measures how route preparation, entry conditioning, and
the frozen locomotion actors together respond to the change in geometry. We report this cohort
separately from the 2,800-run comparison.}

\begin{table}[ht]
\centering\small
\begin{tabular}{@{}llrrrr@{}}
\toprule
Scale & Task & Prior & Live & Shortest & RB-TRG\\
\midrule
0.8 & wp05\_08\_stairs & 2 & 5 & 7 & 1\\
0.8 & wp17\_21\_platform & 7 & 2 & 4 & 2\\
0.8 & wp26\_29\_terrace & 1 & 1 & 5 & 6\\
0.8 & wp28\_30\_descent & 3 & 7 & 6 & 9\\
1.0 & wp05\_08\_stairs & 3 & 5 & 5 & 1\\
1.0 & wp17\_21\_platform & 5 & 2 & 4 & 5\\
1.0 & wp26\_29\_terrace & 1 & 4 & 6 & 5\\
1.0 & wp28\_30\_descent & 4 & 3 & 4 & 4\\
1.1 & wp05\_08\_stairs & 3 & 3 & 6 & 1\\
1.1 & wp17\_21\_platform & 2 & 2 & 1 & 1\\
1.1 & wp26\_29\_terrace & 0 & 1 & 0 & 0\\
1.1 & wp28\_30\_descent & 1 & 3 & 3 & 4\\
1.2 & wp05\_08\_stairs & 3 & 1 & 4 & 0\\
1.2 & wp17\_21\_platform & 2 & 4 & 4 & 1\\
1.2 & wp26\_29\_terrace & 0 & 0 & 1 & 0\\
1.2 & wp28\_30\_descent & 4 & 1 & 2 & 4\\
\bottomrule
\end{tabular}
\caption{\rev{Full ordered-completion counts for the geometry cohort. Each cell has ten
attempts, all under pose noise. Tasks and paired starts are fixed within each scale. The 1.0
setting is the unscaled control on the same host.}}
\label{tab:geometrycounts}
\end{table}
\begin{figure}[t]
\centering
\includegraphics[width=0.72\linewidth]{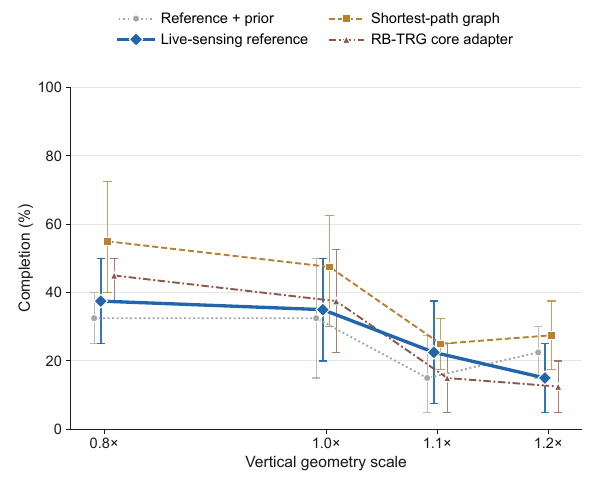}
\caption{\rev{\textbf{Completion under vertical scaling of the scene.} Each point combines four tasks and ten paired pose-noise seeds (40 trials per
method and scale). Error bars are 95\% percentile seed-block bootstrap intervals from 20,000
draws, which keep all four tasks together within each seed. The markers are offset
horizontally only for visibility; all four methods use exactly the scale shown. Lines connect
the measured settings; intermediate scales were not evaluated. Vertical scaling changes one
scene, and robot parameters stay fixed. The descent task includes an approach waypoint and
differs from the earlier multi-terrain descent subroute.}}
\label{fig:geometry-boundary}
\end{figure}

\begin{figure}[t]
\centering
\includegraphics[width=\linewidth]{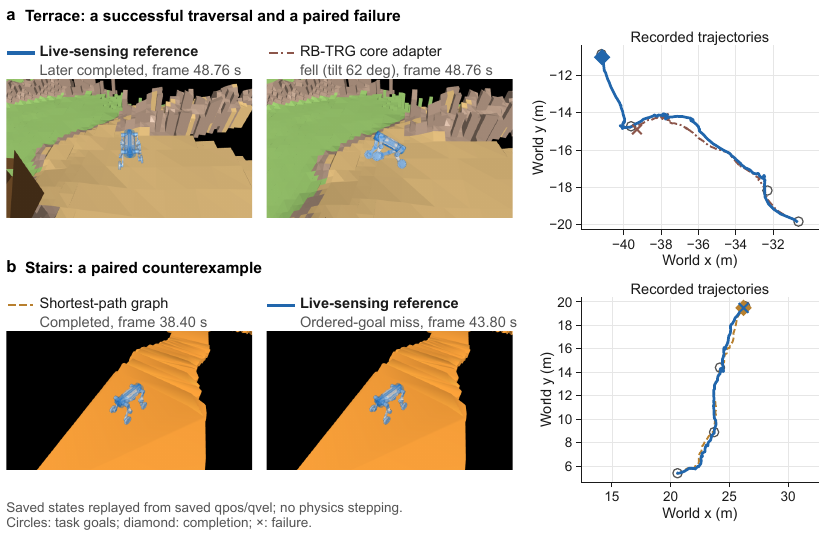}
\caption{\rev{\textbf{Matched replays of two discordant trial pairs.} The
frames and full XY traces come from saved states of matched pose-noise trials: terrace seed
3007 and stairs seed 3012. For each task, we selected the lower-median successful completion
time among the discordant pairs of the further 30-seed cohort (two terrace candidates and six
stair candidates). The selected cases show discordant outcomes. On stairs, the live-sensing run
ends with software completion but misses the ordered-goal criterion, so it is scored as a
failure. The terrace frames use the matched failure time of 48.76 s; the live-sensing run
completes later, at 61.50 s. The stair frames show their own labelled terminal times. The
replays load the recorded qpos/qvel and compute forward kinematics at each saved state. Both
methods use the same camera orientation and distance, centred on their recorded robot
positions.}}
\label{fig:paired-simulation-replays}
\end{figure}

\end{document}